\documentclass[Afour,sageh,times]{sagej}
\usepackage{moreverb,url}
\usepackage[colorlinks,bookmarksopen,bookmarksnumbered,citecolor=red,urlcolor=red]{hyperref}
\usepackage[normalem]{ulem}
\usepackage{subfig}
\usepackage{graphicx}
\useunder{\uline}{\ul}{}
\usepackage{appendix}
\usepackage{booktabs}
\usepackage{pifont}
\usepackage{siunitx}
\usepackage{lscape}
\usepackage{titletoc} 
\usepackage{amsmath}
\usepackage{gensymb}
\usepackage{textcomp}
\usepackage{pdflscape} 
\usepackage{xcolor} 
\usepackage{tikz}
\usetikzlibrary{positioning, spy}

\newcommand{\equalcontribution}{\textsuperscript{*}}

\begin{document}


\title{
THÖR-MAGNI: A Large-scale Indoor Motion Capture Recording of Human Movement and Robot Interaction}

\author{Tim Schreiter\equalcontribution\affilnum{1}, Tiago Rodrigues de Almeida\equalcontribution\affilnum{1}, Yufei Zhu\affilnum{1}, Eduardo Gutierrez Maestro\affilnum{1}, Lucas Morillo-Mendez\affilnum{6}, Andrey Rudenko\affilnum{2}, Luigi Palmieri\affilnum{2},  Tomasz P. Kucner\affilnum{3,4}, Martin Magnusson\affilnum{1}, Achim J. Lilienthal\affilnum{1,5}}

\affiliation{\affilnum{1}\"{O}rebro University, Sweden\ \affilnum{2}Robert Bosch GmbH, Corporate Research, Stuttgart,
Germany.\ \affilnum{3}Finnish Center for Artificial Intelligence (FCAI), Finland.\ \affilnum{4}School of Electrical Engineering Aalto University.\ \affilnum{5} Technical University of Munich, Germany.\ \affilnum{6} Independent Researcher, Spain.\ \affilnum{$^{*}$}These authors contributed equally to the work.}

\corrauth{Tim Schreiter}
\email{tim.schreiter@oru.se}
   

\begin{abstract}
We present a new large dataset of indoor human and robot navigation and interaction, called THÖR-MAGNI, that is designed to facilitate research on social navigation: e.g., modelling and predicting human motion, analyzing goal-oriented interactions between humans and robots, and investigating visual attention in a social interaction context.
THÖR-MAGNI was created to fill a gap in available datasets for human motion analysis and HRI. This gap is characterized by a lack of comprehensive inclusion of exogenous factors and essential target agent cues, which hinders the development of robust models capable of capturing the relationship between contextual cues and human behavior in different scenarios.
Unlike existing datasets, THÖR-MAGNI includes a broader set of contextual features and offers multiple scenario variations to facilitate factor isolation. The dataset includes many social human-human and human-robot interaction scenarios, rich context annotations, and multi-modal data, such as walking trajectories, gaze tracking data, and lidar and camera streams recorded from a mobile robot.
We also provide a set of tools for visualization and processing of the recorded data.
THÖR-MAGNI is, to the best of our knowledge, unique in the amount and diversity of sensor data collected in a contextualized and socially dynamic environment, capturing natural human-robot interactions.
\end{abstract}
\maketitle

\section{Introduction}
In recent years, the topics of human motion modeling, prediction and interaction with social and service robots have been rapidly growing, driven by industrial interests and a quest for safer algorithms in human-robot interaction settings. Various types of advanced automated systems, such as mobile robots (including autonomous vehicles), manipulators, and sensor networks, benefit from human motion models for safe and efficient operation in the presence of humans. Human motion data is central to human-aware path planning, collision avoidance, tracking, interaction, understanding human activities, and collaborating on shared tasks.

Modern approaches for modeling human motion require plentiful data recorded in diverse environments and settings to train on, as well as for the evaluation \citep{rudenko_survey_20}. Among the growing numbers of human trajectory datasets, most focus on capturing interactions between the moving agents in indoor \citep{atc_13}, outdoor \citep{sdd_16} and automated driving \citep{bock2020ind} settings. These datasets are designed to study how people interact and avoid collisions in social settings, by describing their motion through position and velocity information. Further datasets attempt to capture  full-body motion in various activities and human-object interactions in household settings \citep{liu2019ntu,mogaze_20,jrdb_act22}.

Human motion is influenced by many exogenous factors, which cumulatively amount to the \emph{context} in which people move and interact. Among those are numerous
environment factors: motion and activities of other people and robots, locations of obstacles, semantic attributes such as points of common interest, direction signs and special zones. Motion datasets should not only capture these factors to enable computational analysis of how people navigate, but also vary them systematically to support factor isolation in various conditions. Datasets with access to rich context can help to better explain, model and predict human motion.

Furthermore, beyond the environment context, there are various aspects of the specific person --- \emph{target agent cues}~\citep{rudenko_survey_20} --- which are helpful in better understanding their intention, ongoing activity, attention and distraction, preferences and abilities. These cues include head orientations and full body positions, gaze directions, social grouping and past activity patterns. Multi-modal approaches for human motion modeling and prediction can provide more accurate results by combining these cues~\citep{de_Almeida_2023_ICCV}, and their development is subject to the availability of high-quality multi-modal data.

\begin{figure*}[!t]
    \centering
    \includegraphics*[width=\textwidth]{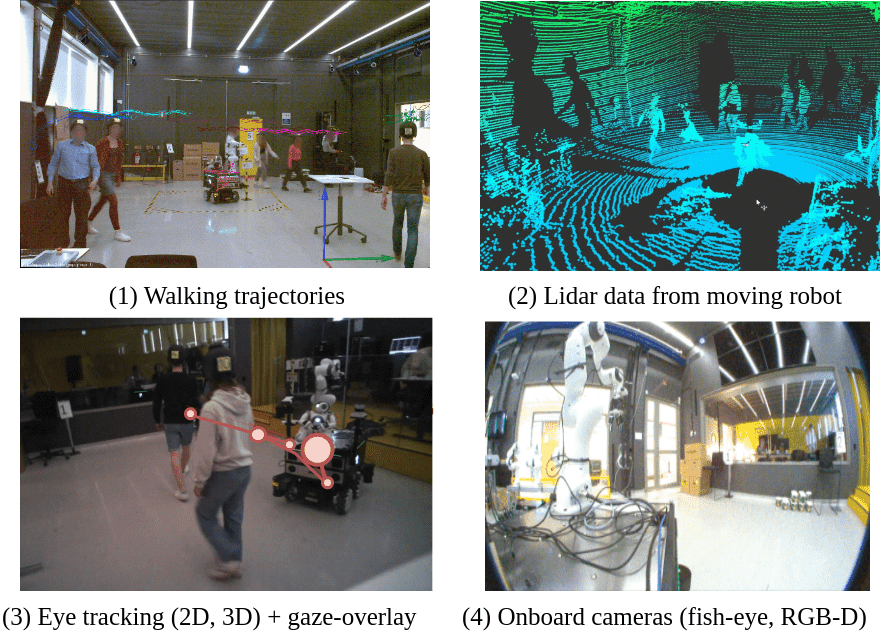}
    \caption{TH\"{O}R-MAGNI data modalities. (1) walking trajectories of participants, in a workplace setting shared with other humans and robots; (2) lidar sweep recorded with a mobile robot; (3) snapshot from an eye tracker's gaze overlay video; (4) fish-eye camera image from the mobile robot, showing object stashes and two goal points from our scenarios.}
    \label{fig:thor_MAGNI_data}
\end{figure*}

Existing datasets in the field of human motion analysis often lack the comprehensive inclusion of the exogenous factors and the target agent cues necessary for holistic studies of human motion dynamics. This research gap hinders the development of robust models capable of capturing the relationship between contextual cues and human behavior in different scenarios. To address this gap, we present a novel dataset that not only incorporates a broader set of contextual features, but also contains multiple variations to support factor isolation. By integrating diverse modalities such as walking trajectories, eye tracking data, and environmental sensory inputs captured by a mobile robot (see Figure~\ref{fig:thor_MAGNI_data}), our dataset fosters the exploration and analysis of human motion in various scenarios with increased fidelity and granularity.
In this paper, we propose a novel dataset of accurate human and robot navigation and interaction in diverse indoor contexts, building on the previous TH\"{O}R dataset \citep{thor_20}. The TH\"{O}R dataset pioneered weakly-scripted scenario-based data collection with motion capture in a controlled environment, recording continuous activities, which involve meaningful social navigation towards randomized targets in the environment. Our new TH\"{O}R-MAGNI dataset extends this effort with rich context annotations, time-synchronized multi-modal data, human-robot interaction scenarios and diverse navigation modes of a mobile robot.

The TH\"{O}R-MAGNI data collection is designed around systematic variation of factors in the environment to allow building cue-conditioned models of human motion and verifying hypotheses on factor impact. To that end, we propose several scenarios in which the participants, in addition to basic navigation, need to move objects, interact with each other and the robot, and respond to remote instructions. The dataset includes differential and omnidirectional robot navigation, semantic zones and direction signs in the environment, and many further aspects. We provide position and head orientation for each moving agent, 3D lidar scans and gaze tracking. Finally, we provide tools to visualize the multiple modalities of the dataset and to preprocess the trajectory data. In total, TH\"{O}R-MAGNI captures 3.5 hours of motion of 40 participants over 5 days of recording, which is available for download\footnote{\url{https://doi.org/10.5281/zenodo.10407223}}. Furthermore, we note the continuity between the TH\"{O}R and TH\"{O}R-MAGNI recordings due to their shared environment (in diverse configurations), motion capture system and complimentary scenario composition.

In this paper, we motivate and detail the TH\"{O}R-MAGNI data collection and sensor setup, describe the interfaces to the dataset and compare it to the prior datasets. The paper is structured as follows: in Section~\ref{sec:related} we review the prior state-of-the-art datasets and in Section~\ref{sec:context} we outline the target application domains. Section~\ref{sec:dataset} provides all necessary information about the data collection, and Section~\ref{sec:usage} describes the data formats and tools used to visualize and preprocess the data. Finally, Section~\ref{sec:results} presents a quantitative evaluation of the collected data followed by a conclusion in Section~\ref{sec:conclusions}.



\section{Related Work}\label{sec:related}

Multi-modal human motion datasets, including gait patterns, gaze vectors, human-robot interactions, and robot sensor data drive a wide range of research applications.
These include human motion prediction~\citep{rudenko_survey_20, kothari_21}, human motion
representation for mobile robots~\citep{tomek_23}, 
human-robot interaction~\citep{survey_hri_23}, human awareness in robot motion planning~\citep{faroni22,heuer2023proactive}, and gaze-based 
prediction of human pose and locomotion mode~\citep{gimo_22, gaze_locomotion_22}.

Early datasets such as UCY~\citep{ucy_07} and ETH~\citep{eth_09} have contributed significantly to our understanding of human movement in outdoor 
environments. 
Although these datasets encompass a range of human motion attributes such as trajectories, group identification, and goal points, social interactions play minor role shaping the human trajectories~\citep{Makansietal21}.
The indoor ATC dataset introduced by~\citet{atc_13} represents a data collection with high coverage and tracking accuracy due to the use of 49 range sensors for raw data acquisition. The tracking method involved the independent estimation of positions and body orientations from each sensor, which were subsequently fused together. This fusion process increased the robustness of the primary estimates and ensured a high degree of accuracy in the resulting dataset. In contrast to the UCY and ETH datasets, our dataset contains many social interactions, as we always had multiple participants moving in the same space between goal points, which deliberately allowed for frequent interactions between participants (see Section~\ref{goal_driven}). Furthermore, unlike the ATC dataset, we have included a mobile robot in the scene, which allows the study of human-robot interaction scenarios (see Section~\ref{sHRI}).

\cite{ktp_14}, \cite{kth_15}, and \cite{jrdb_act22} have presented human motion datasets acquired through mobile robotic systems. While the datasets presented by~\cite{ktp_14} and \cite{kth_15} consist of short acquisitions and have limited contextual information such as maps or environmental goals, \cite{jrdb_act22} have contributed a more comprehensive dataset. Their dataset includes detailed annotations of micro-actions and social group dynamics, thereby offering a richer and more contextualized understanding of human motion patterns in diverse environments.
However, in these datasets,  
human locations are based on detections in the sensor's field of view onboard the mobile robot, which limits the scope of tracking due to occlusions. In contrast to these works, we used a motion capture system to track the moving agents (described in Section~\ref{subsec:system}), which provides longer
continuous tracking of each observed agent.

\cite{mogaze_20} presented the MoGaze dataset, a notable advancement by incorporating a motion capture system for 
full-body pose tracking and eye-tracking data for humans engaged in various activities. Similarly, \cite{chen2022human} proposed a human tracking dataset for recording human-robot cooperation tasks in retail environments.
However, both datasets do not capture social interactions as they track only one person. In addition, MoGaze does not include a mobile robot in the scene. 
The absence of these elements hinders the study of downstream applications, for instance robot motion planning methods in the ``invisible robot'' settings \citep{heuer2023proactive}, in which the humans do not react to the robot's motion and location, but rather the full extent of collision avoidance falls on the robot. 
Similar to TH\"{O}R-MAGNI, the  TH\"{O}R dataset introduced by \cite{thor_20} presents accurate human motion trajectories in the presence of a robot. 
While the TH\"{O}R dataset provides tracking accuracy in a socially dynamic environment, its limited recording duration (1 hour) poses 
challenges for in-depth studies, particularly concerning data-intensive deep learning-based methods for trajectory 
prediction.

In summary, the TH\"{O}R-MAGNI dataset based on the protocol proposed by~\cite{thor_20}, overcomes the limitations of its predecessors.
Table~\ref{tab:datasets} shows a thorough comparison with well-established and recent datasets.
TH\"{O}R-MAGNI contains 3.5 times more trajectory data than TH\"{O}R, therefore providing a broader range of situations for 
the analysis of human motion trajectories. In addition, TH\"{O}R-MAGNI includes sensor data recorded by a mobile robot. Furthermore, our dataset provides gaze vectors aligned with the corresponding 
trajectories, giving the opportunity to simultaneously analyze both modalities. 
This alignment not only enables studies of human-robot interaction but also facilitates 
in-depth analyses of the complex interplay between human visual attention and motion patterns. Finally, TH\"{O}R-MAGNI is, to the best of our knowledge, unique in the amount and diversity of sensor data collected in a contextualized and socially dynamic environment, capturing natural human-robot interactions.
\begin{landscape}
\begin{table}[t]
    \centering
    \caption{Comparison of human trajectory datasets.}
    \label{tab:datasets}
    \resizebox{\columnwidth}{!}{%
    \begin{tabular}{@{}c|c|c|c|c|c|c|c|c|c|c|c|c@{}}
    \toprule
      \textbf{Dataset}&
      \textbf{Environment} &
      \begin{tabular}[c]{@{}c@{}}\textbf{Sensors for Pose}\\ \textbf{Estimation}\end{tabular} &
      \textbf{Duration} &
      \begin{tabular}[c]{@{}c@{}}\textbf{Pose}\\ \textbf{Frequency (Hz)}\end{tabular} &
      \begin{tabular}[c]{@{}c@{}}\textbf{Pose}\\ \textbf{Annotation}\end{tabular} &
      \begin{tabular}[c]{@{}c@{}}\textbf{Social} \\ \textbf{Interactions}\end{tabular} &
      \begin{tabular}[c]{@{}c@{}}\textbf{Robot in the} \\ \textbf{Scene}\end{tabular} &
      \begin{tabular}[c]{@{}c@{}}\textbf{Intended} \\ \textbf{for HRI}\end{tabular} &
      \textbf{Goals} &
      \textbf{Map} &
      \textbf{Robot Data} &
      \textbf{Other Data} \\ \midrule
    \begin{tabular}[c]{@{}c@{}}UCY\\ \citep{ucy_07}\end{tabular} &
    \begin{tabular}[c]{@{}c@{}}Street\\ (outdoor)\end{tabular} &
      RGB camera &
      20 min. &
      Continuous &
      Manual &
      \ding{51} &
        &
        &
        &
        &
        &
        \\ \midrule
    \begin{tabular}[c]{@{}c@{}}ETH\\ \citep{eth_09}\end{tabular} &
      University and Hotel &
      RGB camera &
      25 min. &
      2.5 &
      Manual &
      \ding{51} &
        &
        &
      \ding{51} &
      \ding{51} &
        &
        \\ \midrule
    \begin{tabular}[c]{@{}c@{}}Edinburgh\\ \citep{edinburgh_09}\end{tabular} &
      \begin{tabular}[c]{@{}c@{}}Forum\\ (outdoor)\end{tabular} &
      RGB camera &
      ~4 months &
      6-10 &
      Automated &
      \ding{51} &
        &
        &
        &
        &
        &
       \\ \midrule
    \begin{tabular}[c]{@{}c@{}}Town Center\\ \citep{town_center_11}\end{tabular} &
    \begin{tabular}[c]{@{}c@{}}Street\\ (outdoor)\end{tabular} &
      RGB camera &
     5 min. &
      25 &
      Manual &
      \ding{51} &
        &
        &
        &
      Raw &
        &
        \\ \midrule
    \begin{tabular}[c]{@{}c@{}}VIRAT\\ \citep{virat_11}\end{tabular} &
      Various outdoors &
      RGB camera &
      29 h &
      2, 5, 10 &
      Manual &
      \ding{51} &
        &
        &
        &
      Raw &
        &
      \begin{tabular}[c]{@{}c@{}}Human activities, \\ agents types\end{tabular} \\ \midrule
    \begin{tabular}[c]{@{}c@{}}Central station\\ \citep{central_station_12}\end{tabular} &
      Train station &
      RGB camera &
      34 min. &
      24 &
      Automated &
      \ding{51} &
        &
        &
        &
        &
        &
        \\ \midrule
    \begin{tabular}[c]{@{}c@{}}ATC \\ \citep{atc_13}\end{tabular} &
      Shopping Centre &
      \begin{tabular}[c]{@{}c@{}}Several 3D\\ range sensors\end{tabular} &
      41 days &
      10-30 &
      Automatic &
      \ding{51} &
        &
        &
        &
        &
        &
        \\ \midrule
      \begin{tabular}[c]{@{}c@{}}NBA SportVU\\ 2013~\footnote{\url{https://github.com/linouk23/NBA-Player-Movements}}\end{tabular} &
      Basketball court &
      RGB camera &
      20 days &
      25 &
      Automatic &
        &
        &
        &
        &
        &
        &
      \begin{tabular}[c]{@{}c@{}}Multi-agent human\\ activities\end{tabular} \\ \midrule
    \begin{tabular}[c]{@{}c@{}}KTP \\ \citep{ktp_14}\end{tabular} &
      Empty Room &
      RGB-D camera &
      4.7 min. &
      30 &
      Manual &
      \ding{51} &
      \ding{51} &
        &
        &
        &
      RGB-D camera &
      Motion capture \\ \midrule
    \begin{tabular}[c]{@{}c@{}}KTH\\ \citep{kth_15}\end{tabular} &
      Lab &
      \begin{tabular}[c]{@{}c@{}}RGB-D camera and\\ 2D laser scanner\end{tabular} &
      2.7 h &
      25 &
      Automatic &
      \ding{51} &
      \ding{51} &
        &
        &
        &
      \begin{tabular}[c]{@{}c@{}}RGB-D camera and\\ 2D laser scanner\end{tabular} &
        \\ \midrule
    \begin{tabular}[c]{@{}c@{}}UCLA Aerial Event Dataset\\ \citep{ucla_15}\end{tabular} &
      Outdoor spaces &
      RGB camera &
      1.5 h &
      60 &
      Automatic &
      \ding{51} &
        &
        &
        &
      Raw &
        &
      \begin{tabular}[c]{@{}c@{}}Human roles, small and\\ large objects location\end{tabular} \\ \midrule
    \begin{tabular}[c]{@{}c@{}}SDD\\ \citep{sdd_16}\end{tabular} &
      \begin{tabular}[c]{@{}c@{}}University campus\\ (outdoor)\end{tabular} &
      RGB camera &
      5 h &
      30 &
      Manual &
      \ding{51} &
        &
        &
        &
      Raw &
        &
      Human activities \\ \midrule
    \begin{tabular}[c]{@{}c@{}}L-CAS\\ \citep{lcas_17}\end{tabular} &
      Office &
      3D lidar &
      49 min. &
      10 &
      Manual &
      \ding{51} &
      \ding{51} &
        &
        &
        &
      3D lidar &
      \begin{tabular}[c]{@{}c@{}}Single-person, group\\ labels\end{tabular} \\ \midrule
    \begin{tabular}[c]{@{}c@{}}MoGaze\\ \citep{mogaze_20}\end{tabular} &
      Lab &
      Motion capture &
      3 h &
      120 &
      Ground truth &
        &
        &
        &
        &
        &
        &
      Human activities \\ \midrule
    \begin{tabular}[c]{@{}c@{}}Flobot\\ \citep{flobot_20}\end{tabular} &
      \begin{tabular}[c]{@{}c@{}}Public spaces (i.e., airport, \\ warehouse, supermarket)\end{tabular} &
      \begin{tabular}[c]{@{}c@{}}3D lidar and\\RGB-D camera\end{tabular} &
      27.5 min. &
      10 &
      Automatic &
      \ding{51} &
      \ding{51} &
        &
        &
        &
      \begin{tabular}[c]{@{}c@{}}2D and 3D lidars, RGB-D\\ and stereo cameras\end{tabular} &
        \\ \midrule
    \begin{tabular}[c]{@{}c@{}}TH\"{O}R\\ \citep{thor_20}\end{tabular} &
      \begin{tabular}[c]{@{}c@{}}Lab with various\\ spatial layouts\end{tabular} &
      Motion capture &
      1 h &
      100 &
      Ground truth &
      \ding{51} &
      \ding{51} &
        &
      \ding{51} &
      \ding{51} &
      3D lidar &
      \begin{tabular}[c]{@{}c@{}}Aligned ET,\\ Human activities\end{tabular} \\ \midrule
      \begin{tabular}[c]{@{}c@{}}JRDB-Act\\ \citep{jrdb_act22}\end{tabular} &
        \begin{tabular}[c]{@{}c@{}}University campus\\ (indoor and outdoor)\end{tabular} &
        \begin{tabular}[c]{@{}c@{}}Lidar and\\RGB camera\end{tabular} &
        1 h &
        7.5 &
        Automatic &
        \ding{51} &
        \ding{51} &
          &
          &
          &
        \begin{tabular}[c]{@{}c@{}}Velodyne, several cameras\\ (RGB and RGB-D)\end{tabular} &
        Human activities \\ \midrule
    \begin{tabular}[c]{@{}c@{}}Oxford-IHM\\ \citep{oxford_23}\end{tabular} &
      Lab/Office &
      Motion capture &
      1 h &
      100 &
      Ground truth &
        &
      \ding{51} &
        &
      \ding{51} &
      \ding{51} &
      RGB-D camera &
      Static RGB-D camera\\ \midrule
    \begin{tabular}[c]{@{}c@{}}\textbf{TH\"{O}R-MAGNI}\\ \textbf{(2024)}\end{tabular} &
      \begin{tabular}[c]{@{}c@{}}Lab with various\\ spatial layouts\end{tabular} &
      Motion capture &
      3.5 h &
      100 &
      Ground truth &
      \ding{51} &
      \ding{51} &
      \ding{51} &
      \ding{51} &
      \ding{51} &
      \begin{tabular}[c]{@{}c@{}} 3D lidar, RGB and\\ RGB-D cameras\end{tabular} &
      \begin{tabular}[c]{@{}c@{}}Aligned ET, \\ Several human \\ activities\end{tabular} \\ \bottomrule
    \end{tabular}%
    }
    \end{table}
\end{landscape}
\section{Context of the TH\"{O}R-MAGNI Dataset} \label{sec:context}

The THÖR-MAGNI dataset provides diverse navigation styles of a mobile robot and humans engaged in various activities in a shared environment with robotic agents, and incorporates multi-modal data for a more complete representation. Following a comparative analysis of our dataset with state-of-the-art datasets contributing to the evolving landscape of human motion research (see Section~\ref{sec:related}), this section supports users of our dataset by providing a detailed exploration of its features in the context of human motion and robot navigation and interactions. We explain their significance in addressing identified gaps, before describing the dataset itself in Section~\ref{sec:dataset}. 

\subsection{Goal-directed Human Motion Trajectories}\label{goal_driven}
The presence of goal-directed human agents is crucial in the field of human motion 
prediction~\citep{dendorfer21,Zhao_2021_ICCV,Chiara_2022_CVPR}.
Traditional approaches often depict human agents as rational entities, acting logically and moving towards specific goals or destinations \citep{ziebart2009planning}. Real-world recordings commonly show this directional traffic flow, characterized by distinct goal points, often resulting in a consistent and linear motion with limited diversity. In our dataset, we include scenes with seven distinct goal points distributed over a larger spatial volume, and scenes where they are arranged in a more compact space (see Section \ref{sec:dataset}). Goal points and static obstacles are positioned strategically to ensure that recorded trajectories are not only sufficiently long but also topologically diverse, i.e. covering a range of spatial arrangements and configurations. This approach allows for the inclusion of frequent interactions between the moving agents, contributing to a more comprehensive understanding of human motion dynamics.

\subsection{Navigation of Heterogeneous Agents}\label{nav_human}

Heterogeneous agents are dynamic entities that navigate with distinct motion patterns. This heterogeneity stems from various factors that affect the motion such as tasks and ongoing activities performed by the agent~\citep{de_Almeida_2023_ICCV}. 
For instance,
several works have studied how humans move individually or as part of a social group~\citep{moussaid2010walking,rudenko2018human,wang2022group}. 
It has been shown that humans can coordinate their movements 
as a group by following simple rules based on the visual perception of 
local motion~\citep{boos2014leadership}. Previous research on the anatomy of leadership in 
collective behavior \citep{garland2018anatomy} describes human 
collective behavior as optimal coordination and leadership dynamics in various group scenarios. In particular, crowd dynamics are 
not only determined by physical constraints, but also significantly influenced by 
communicative and social interactions among individuals~\citep{moussaid2010walking}. Autonomous driving datasets often highlight the motion of heterogeneous agents in mixed traffic \citep{chandra19,salzmann20}. In our dataset, we introduce roles for participants 
tailored for industrial tasks, such as navigating alone or in groups of different sizes, transporting 
various objects and interacting with a robot. This heterogeneous social setting provides a novel way to study how specific industrial roles 
influence human motion, aligning with the work conducted by~\cite{de_Almeida_2023_ICCV}.

\subsection{Navigation of a Robotic Agent}\label{nav_robot}
Human-aware robot motion planning is crucial for safe navigation in shared spaces, 
especially in the narrow and crowded indoor environments \citep{cancelli2023exploiting}. Understanding human interaction with 
robots of different driving styles promotes the design of socially acceptable motion 
planners~\citep{moller2021survey}. Analyzing participant behavior with robots of varied 
movement patterns reveals insights into how robot motion style affects human 
expectations~\citep{karnan2022socially}, guiding the development of robots that interact 
safely and are well-received by people \citep{shah2023gnm}. Our dataset features scenarios with a mobile robot in teleoperated and semi-autonomous modes and two driving styles: differential drive (forward, backward and turning) and omnidirectional mode (allowing the robot to drive in any direction while keeping its heading). This variety of motion modes (detailed in Section~\ref{robot_mode}) extends the state-of-the-art datasets of teleoperated navigation which feature a single driving style \citep{karnan2022socially}.

\subsection{Spatial Human-Robot Interaction in Shared Workplace Settings}\label{sHRI}
The concept of Industry 5.0 aims to prioritize human well-being in manufacturing 
systems~\citep{leng2022industry}. This requires enhancing the quality of human-machine and 
human-robot interactions in these environments. Designing robots that can clearly express their intentions to human collaborators is a crucial step towards fostering mutual understanding and enhancing the well-being of human workers who regularly interact with robots~\citep{pascher2023communicate}. Furthermore, intuitive human-robot interaction (HRI) not only improves well-being but also enhances safety and efficiency in collaborative settings~\citep{haddadin2011towards}.

Spatial HRI (sHRI) and navigation in shared environments are research areas that have an 
adherent need for accurate datasets of human motion tracking and prediction~\citep{thor_20, chen2022human} and for robots that understand the underlying physical 
interactions between nearby agents and objects~\citep{castri2022causal}.  Our dataset contains recordings of explicit interactions between a mobile robot and individuals in shared workplace settings. TH\"{O}R-MAGNI is a valuable resource for studying human responses to robotic approach and assistance initiatives, enabling researchers to analyze goal-oriented interactions between humans and robots.

\subsection{Eye tracking and Head Orientation in Navigation Tasks}\label{human_gaze}
Eye tracking is a powerful method to study various aspects of human behavior, 
including attention, emotion, cognition, and decision-making, with applications spanning 
education, marketing, gaming, and healthcare~\citep{duchowski2017eye}. Eye tracking provides objective 
data about eye movements and positions and enables researchers to quantify 
visual information processing through various metrics~\citep{duchowski2017eye,mahanama2022eye}. 
In HRI applications, human eye-gaze is an important nonverbal 
signal~\citep{admoni2017social}. In our dataset, we align human gaze data with human motion trajectories, providing an opportunity to study human gaze during visual exploration across dynamic tasks, activities, and scenarios.

Complementary to gaze, head orientation provides another essential 
modality of human behavior, closely related to gaze direction and attentional focus. Head orientation plays a vital role 
in joint attention, i.e. the coordination of attention between individuals focusing 
on the same point of interest \citep{tomasello2014joint}. Furthermore, it is valuable for 
detecting interpersonal dynamics in multi-party interactions \citep{stiefelhagen2002tracking}. 
Beyond its social implications, head orientation becomes a predictive indicator of walking motion goals \citep{holman2021watch} and can enhance human motion prediction through vision-based features \citep{salzmann2023robots}. Using a state-of-the-art motion capture system and eye-tracking devices, our dataset provides highly accurate head poses and orientations aligned with the eye tracking data.

\subsection{Semantic Environment Cues}\label{semantics}
Crucial environmental information, conveyed by semantic cues such as doors, stairs, floor markings, and signs, plays an important role in guiding humans and robots within a given space. These cues, combined with obstacle configurations, influence human interactions with the environment, leading to actions like detouring, bypassing, overtaking, and avoiding specific areas. In our dataset, we include different semantic cues like markings on the floor indicating areas to be cautious of the environment or one way passages that limit the flow of motion in one direction. In this way, we enable the exploration of navigation and interactions in semantically-rich environments. For instance, leveraging Maps of Dynamics~\citep{tomek_23} allows the quantification of motion patterns changes around these cues. This information, in turn, can be utilized to predict long-term human motion dynamics, as demonstrated by~\cite{zhu2023clifflhmp}.



\section{Description of the TH\"{O}R-MAGNI Dataset}\label{sec:dataset}

The THÖR-MAGNI dataset is a large-scale indoor motion capture recording of human movement and robot interaction. It consists of 52 four-minute recordings (runs) of participants performing various activities related to navigating alone and in groups, finding and transporting small and large objects, and interacting with robots. THÖR-MAGNI contains over 3.5 hours of motion data for 40 participants, including position, velocity, and head orientation. Eye tracking data is available for 16 of them, totaling 8.3 hours for eight activities (see Table~\ref{tab:recorded_eye_data}). In 24 runs, THÖR-MAGNI also includes the robot sensor data of 3D point clouds from an Ouster lidar. Additionally, videos recorded by an Azure Kinect camera and a Basler fish-eye camera onboard a mobile robot are available on request.


In this section, we detail the environment in which we recorded the data (Section~\ref{subsec:environment}), the navigation and task design for the participants and the robot (Section~\ref{subsec:navigation_interaction}), interactive scenarios to emphasize the various contextual aspects of human motion (Section~\ref{subsec:scenarios}), participants' background and priming (Section~\ref{subsec:prime}) and the technical implementation of the recording pipeline and collection of motion capture and eye tracking data (Section~\ref{subsec:system}).

\subsection{Environment Design}
\label{subsec:environment}

\begin{figure}[!t]
    \centering
    \subfloat{\includegraphics[width=0.48\linewidth]{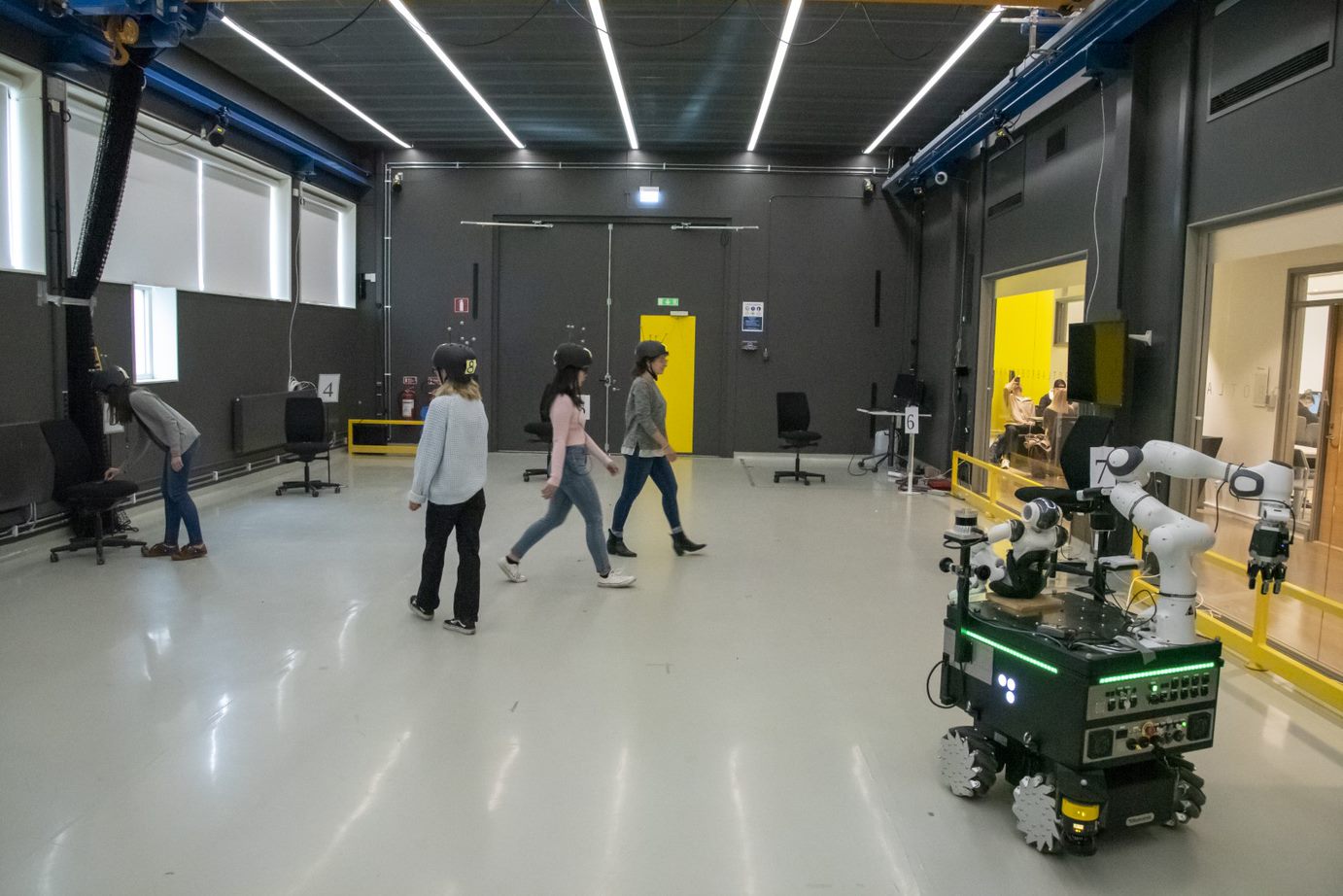}}
    \subfloat{\includegraphics[width=0.48\linewidth]{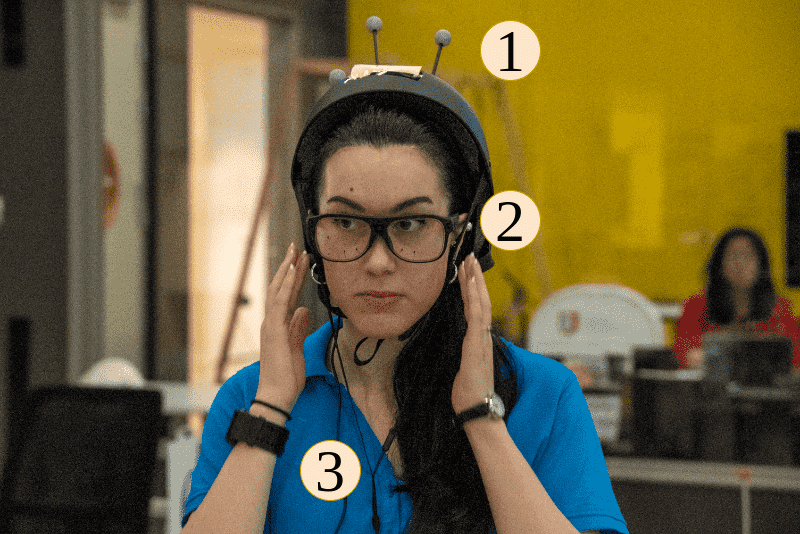}}
    \caption{Our dataset provides a comprehensive exploration of human-robot interaction in a shared workplace environment. \textbf{Left:} Participants navigate independently, collaborate in social groups, and engage with a mobile robot. Navigation between goal points is coordinated via card decks at the goal points that assign a participant a new goal point upon drawing a card, as seen on the far left.
    \textbf{Right:} Equipment utilized in our data collection comprises: (1) bicycle helmets equipped with motion capture tracking markers, (2) eye tracking glasses, and (3) headphones used for receiving spoken instructions.}
    \label{fig:evalFront}
\end{figure}

We conducted the data acquisition in a laboratory at \"{O}rebro University, the same as in the TH\"{O}R dataset~\citep{thor_20}. There are two different configurations for the laboratory. One features a small but free-space environment(see Figure~\ref{fig:evalFront} left). The other resembles an industrial logistics setting and promotes frequent interactions between human and robotic co-workers (see Section ~\ref{subsec:scenarios}). Both room configurations have seven goal positions, to drive purposeful human navigation through the available space, generating frequent interactions in the center. Additionally, we include several environmental layouts (i.e., obstacle maps) in the TH\"{O}R-MAGNI dataset, which vary the placement of static obstacles (robotic manipulators and tables) in the room to prevent walking between goals in a straight path. Apart from static obstacles, two robots are in the room: a static robotic arm near the podium and an omnidirectional mobile robot with a robotic arm on top (see Section~\ref{robot_mode}). 

\subsection{Navigation and Interaction Design}
\label{subsec:navigation_interaction}

The interaction and navigation design in TH\"{O}R-MAGNI extends the weakly-scripted motion recording procedure introduced in the TH\"{O}R dataset \citep{thor_20}. This procedure facilitates realistic motion in controlled settings, in which accurate ground truth motion capture and eye tracking data are collected using specialized equipment (see Figure \ref{fig:evalFront} on the right). Our key idea is to assign meaningful activities and tasks to the recording's participants, allowing them to concentrate on their continuous activity during which they freely move in the room shared with other people and robots. To generate a diverse range of interactions, we developed several scenes that vary in the composition of tasks, robot operation, and other contextual cues, as discussed in Section~\ref{sec:context}.

\subsubsection{Tasks, Activities and Roles Requiring Search and Navigation}\label{subsubsec:activities}

\begin{figure}[!t]
    \centering
    \includegraphics[width=0.9\linewidth]{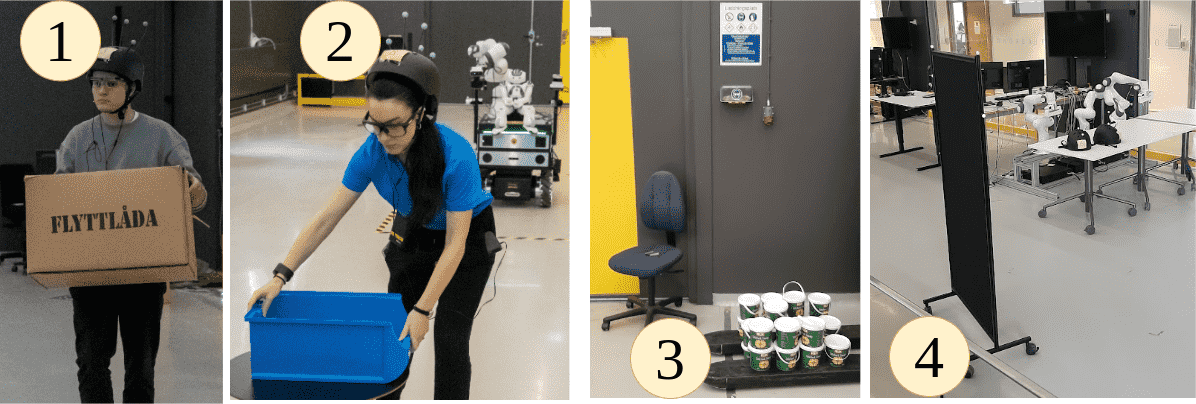}
    \caption{Participants in the role of Carrier were transporting various objects in different size and shapes. (1) {\em Carrier--Box} carrying a medium sized card box, with two hands. (2) {\em Carrier--Storage Bin HRI} placing the bin at a goal point (3) Stash of small objects transported by the {\em Carrier--Bucket} (4) Large Object (poster stand) moved by two {\em Carrier--Large Object}.}
    \label{fig:Carriers}
\end{figure}

We aimed to simulate authentic scenes that reflect the different activities individuals perform in a workplace environment. To that end, we designed several tasks that require search, navigation, interaction with objects, other participants and a mobile robot. Participants engaged in those tasks according to their assigned \emph{role}. 

There are two types of roles in our dataset: \textbf{Visitors} and \textbf{Carriers}. \textbf{Visitors} navigate either individually ({\em Visitors--Alone}) or in groups of two ({\em Visitors--Group 2}) or three {\em Visitors--Group 3}) between target points in the environment. The \textbf{Visitors} role includes a human-robot interaction component denoted by {\em Visitors--Alone HRI}, where participants interact with a robot in a joint navigation task (see Section~\ref{robot_mode}). In addition, \textbf{Carriers} are involved in transporting various objects, including {\em Carrier--Bucket}, {\em Carrier--Box}, {\em Carrier--Storage Bin HRI} and {\em Carrier--Large Object} (see Figure \ref{fig:Carriers}). \textbf{Carriers} transport objects between pre-defined target points, and objects themselves representing different levels of difficulty for navigation, categorized as small (lowest difficulty), medium (medium difficulty), and large (highest difficulty).

\textbf{Visitors} used a card-based system to navigate, receiving new destinations each time they reached a designated goal point. At each goal point, a deck of cards was available, featuring instructions such as ``Go to Goal 1''. The instructions could specify a new destination or contain instructions to go to the robot. In the case of {\em Visitors--Alone}, they drew a card and placed it at the bottom of the deck. Afterward, the participant moved to the destination. In the case of groups, the members could choose who will draw the card.

\textbf{Carriers} were asked to transport objects of different shapes and sizes. These include small objects such as a blue plastic storage bin for the {\em Carrier--Storage Bin HRI} and plastic buckets of canned vegetables for the {\em Carrier--Bucket}, designed for easy, one-handed transportation. For the {\em Carrier--Box}, the participants had to move cardboard boxes as medium-sized objects. These boxes were filled with a few books, allowing for comfortable two-handed transportation. In addition, a collaborative effort involving two participants working as a group featured moving a large object, specifically a poster stand ({\em Carrier--Large Object}). This stand-up, equipped with four wheels, is thin and long and can be moved by two people working in tandem. The overall goal of this setup is to assess how different ongoing activities affect participants' behavioral patterns, including factors such as gaze direction and movement.

\begin{table}
  \caption{Amount of eye tracking- and trajectory data recorded for various activities with all three devices: Tobii~2, Tobii~3 and Pupil Invisible glasses}
  \label{tab:recorded_eye_data}
  \centering
  \resizebox{\columnwidth}{!}{
  \begin{tabular}{{@{}c|c|c@{}}}
    \toprule
    \textbf{Activity}&\begin{tabular}[c]{@{}c@{}}\textbf{Eye tracking (min.)}\end{tabular}&\begin{tabular}[c]{@{}c@{}}\textbf{Trajectory data (min.)}\end{tabular}\\
    \midrule
    {\em Visitors--Alone} & 108 & 392 \\
    \midrule
    {\em Visitors--Group 2} & 124 & 344 \\
    \midrule
    {\em Visitors--Group 3} & 52 & 168 \\
    \midrule
    {\em Visitors--Alone HRI} & 64 & 112 \\
    \midrule
    {\em Carrier--Bucket} & 32 & 96 \\ 
    \midrule
    {\em Carrier--Box} & 60 & 96 \\
    \midrule
    {\em Carrier--Large Object} & 92 & 192 \\
    \midrule
    {\em Carrier--Storage Bin HRI} & 16 & 16 \\ 
    \toprule
  Total & 548 & 1416\\
  \bottomrule
\end{tabular}}
\end{table}

\subsubsection{Modes of Robot Navigation and HRI}\label{robot_mode}

\begin{figure}[!t]
    \centering
    \includegraphics[width=0.7\linewidth]{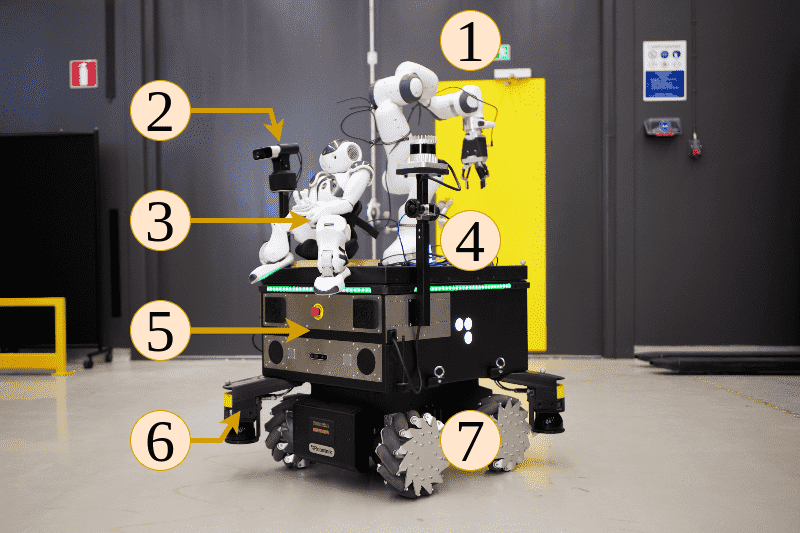}
    \caption{Robot used in- and for data collection (the ``DARKO'' robot) with an omnidirectional mobile base (RB-Kairos) of the dimensions: 760 $\times$ 665 $\times$ 690~mm (5), equipped with two sensor towers, one hosting two Azure Kinect RGB-D cameras (2), and one hosting an Ouster OS0-128 lidar and two Basler fish-eye RGB cameras (4). Additional equipment includes two Sick MicroScan 2D safety lidars (6), mecanum wheels (7), and a NAO robot (``ARMoD'') for interaction with participants (3). The robotic arm with a maximum arm height of 855~mm (1) was not used in our recordings.}
    \label{fig:darko_labeled}
\end{figure}

Our dataset includes a mobile robot, ``DARKO\footnote{\url{https://darko-project.eu/}}'' (see Figure \ref{fig:darko_labeled}), which acts as a static obstacle in some scenes and moves in others. This range of behaviors enables the study of participants' movements and gaze behaviors concerning the stationary and mobile status of the robot. In certain scenes, the robot was teleoperated and moved omnidirectionally, enabling it to reach any 2D position from a stationary position. In some it moved directionally with a predetermined orientation (front). In others the DARKO robot navigated semi-autonomously with manually set goal points. An experimenter was supervising the navigation of DARKO for safety reasons. When acting semi-autonomously the robot interacted with participants through a communication intermediary called the ``Anthropomorphic Robot Mock Driver'' (ARMoD). 

The ARMoD is a small humanoid NAO robot, as shown in Figure \ref{fig:darko_labeled}. It was sitting on the DARKO robot. The ARMoD displayed two  behaviors during interactions: One using only the voice (\textbf{Verbal-Only HRI}). The other uses multi-modal features such as eye contact, robotic gaze, and pointing gestures to support the voice (\textbf{Multi-modal HRI}). This style of interaction reduces fixations on the DARKO robot, increases focus on the ARMoD’s face, and triggers faster response times to instructions of participants, effectively directing attention and improving the quality communication with the robot \citep{schreiter2023advantages}.

\subsection{Scenario Design}\label{subsec:scenarios}

\begin{figure}[!t]
    \centering
    \includegraphics[width=1\linewidth]{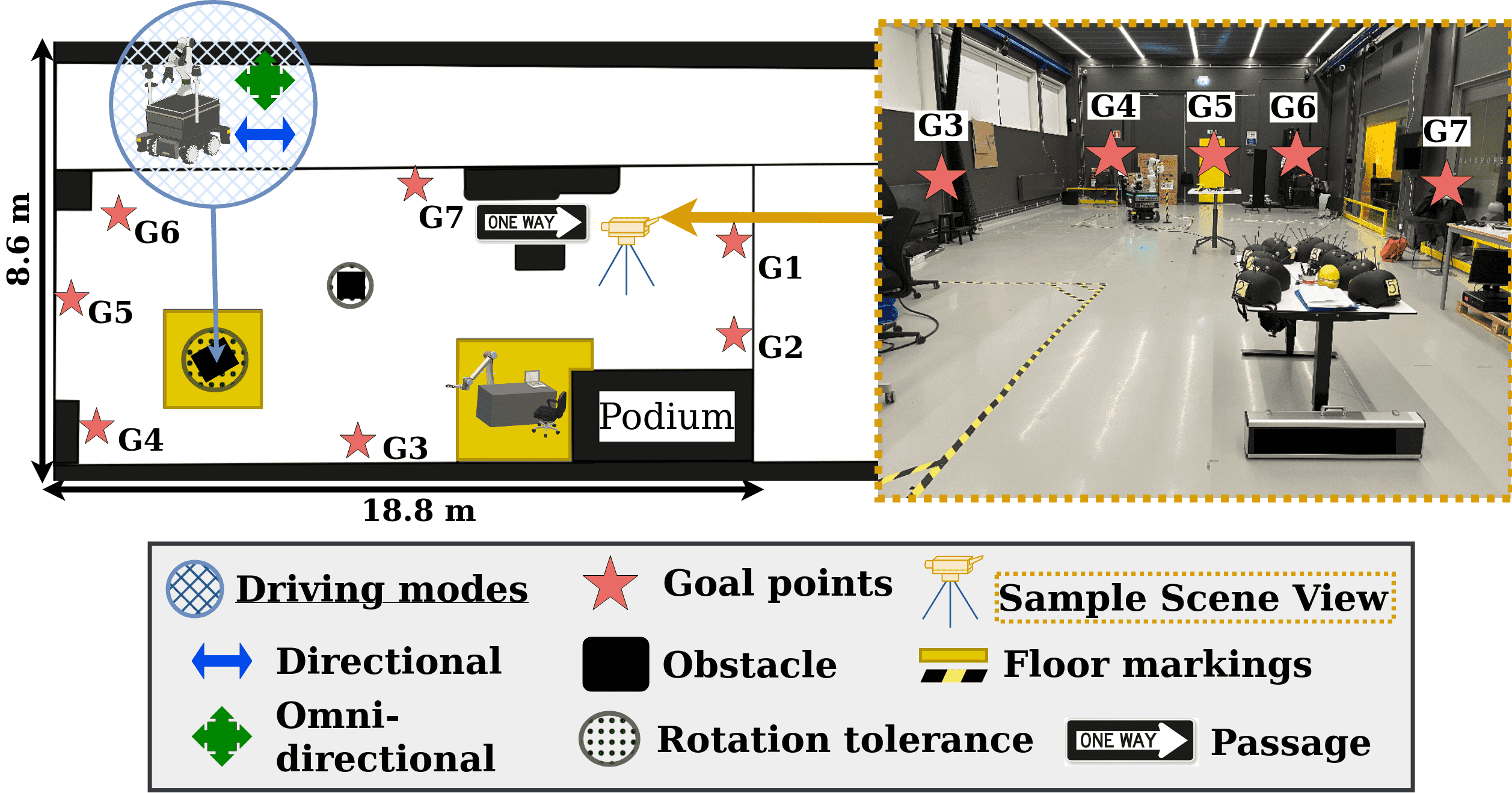} 
    \caption{Varying environmental layouts for the room configuration of Scenarios~1--3. \textbf{Right:} Sample scene view for the site used for data acquisition of the TH\"{O}R-MAGNI dataset showing the room configuration for Scenarios~1--3 with the environment layout for Scenario~1B. \textbf{Left:} Overview of the room configuration and the scenario-specific layout changes. \textbf{Bottom:} Legend explaining elements of the layout, including: Driving styles for the robot in Scenario~3, semantic elements specific for Scenario~1 (Floor markings, Passage) and position of goals and obstacles. Upon placement some objects were subject so a slight rotation between runs, which is accounted for in the layouts with the rotation tolerance.}
    \label{fig:map-scheme}
\end{figure}

We address the context of agent movement by including both humans and robots, as previously discussed, in five specifically designed scenes we call ``scenarios''. Scenario~1 captures the dynamics of motion because of semantic attributes of the environment and sets up a baseline for goal-directed social navigation. Scenario~2 adds role-specific motion for some participants navigating the environment. Subsequently, Scenario~3 explores the impact of different robot motion styles on these role-specific patterns. Figure~\ref{fig:map-scheme} depicts a detailed overview of the room configuration and varying environmental layouts for Scenarios~1--3. Scenario~1's conditions A and B capture regular social behavior in a static environment with and without additional floor markings and a one-way passage. Scenario~2 maintains the same layout as Scenario~1A but introduces individuals performing tasks, emulating industrial activities. Scenario~3 explores human-robot interactions by varying the driving modes of the mobile robot teleoperated by experimenters on a podium.

\begin{figure*}[!t]
\centering
\includegraphics[width=1\linewidth,keepaspectratio]{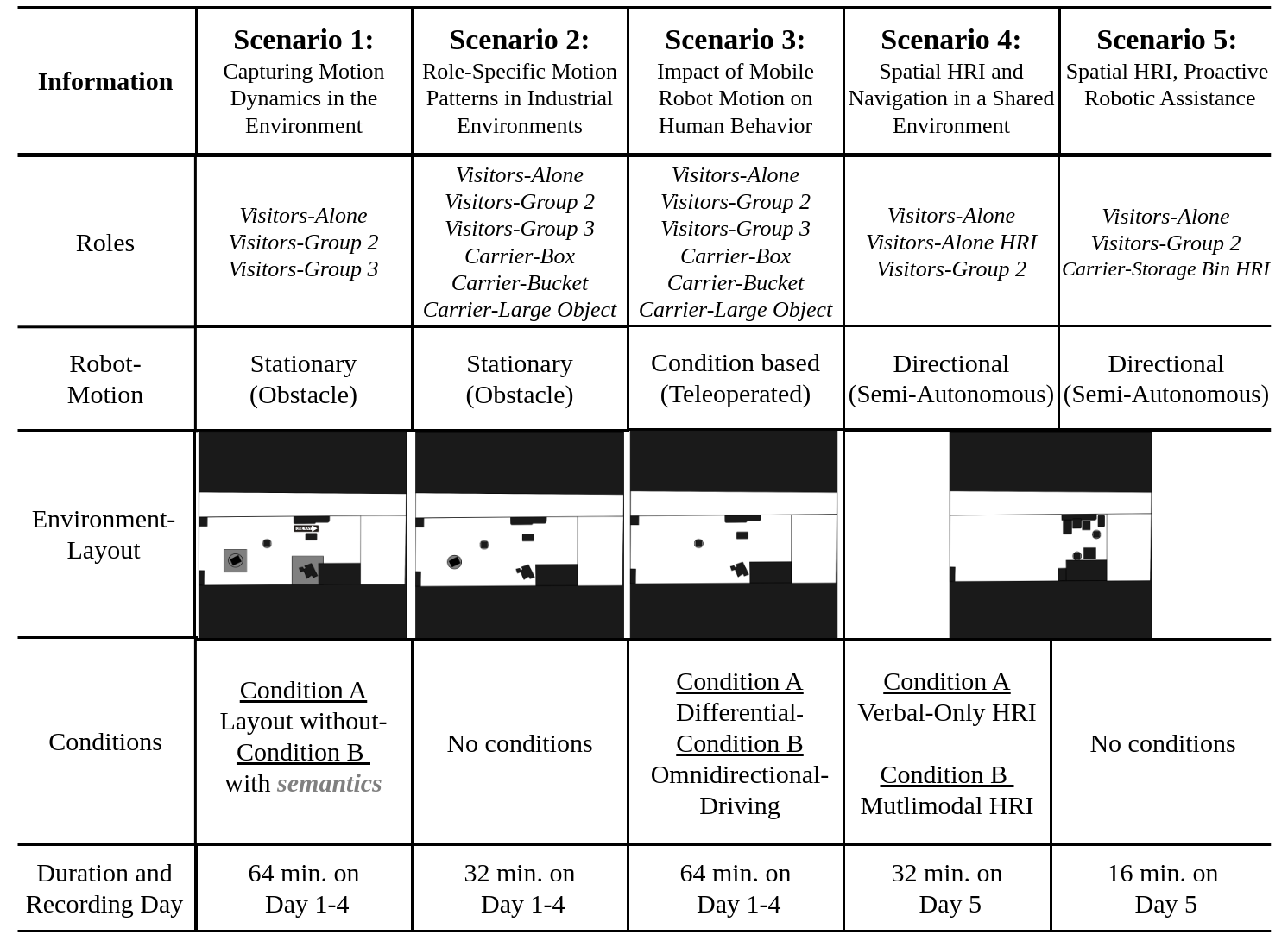}
\caption{Scenario definitions in the TH\"{O}R-MAGNI dataset, including roles, robot motion status (e.g., autonomous or teleoperated), environment layout (i.e., obstacle maps), specific scenario conditions, and duration and recording days.}
\label{fig:Overview_All}
\end{figure*}

Transitioning to a smaller room configuration, we present two scenarios to explore human motion and intended interactions between humans and robots: Scenarios~4 and 5. Scenario~4's participants engaged in intermittent interaction with a mobile robot. This robot communicated in two interaction styles through another entity to mediate joint navigation with participants toward goal points. In Scenario~5, the robots and a human co-worker collaborated actively in transporting small storage bins. For a comprehensive overview of roles and scenarios, see Figure~\ref{fig:Overview_All}.

We recorded multiple runs for each condition in Scenarios~1--5. Specifically, we recorded two runs per condition for Scenarios~1 and 3, two for Scenario~2, four per condition for Scenario~4, and four runs for Scenario~5. To counterbalance learning-based effects, we randomized the recording order of conditions for Scenarios~3 and 5. We implemented this methodical approach to ensure a broad and impartial exploration of the scenarios, capturing subtle interactions and behaviors in each setting.

\subsubsection{Scenario 1: Capturing Motion Dynamics in the Environment}\label{scenario1}

\begin{figure}
    \includegraphics[width=1\linewidth]{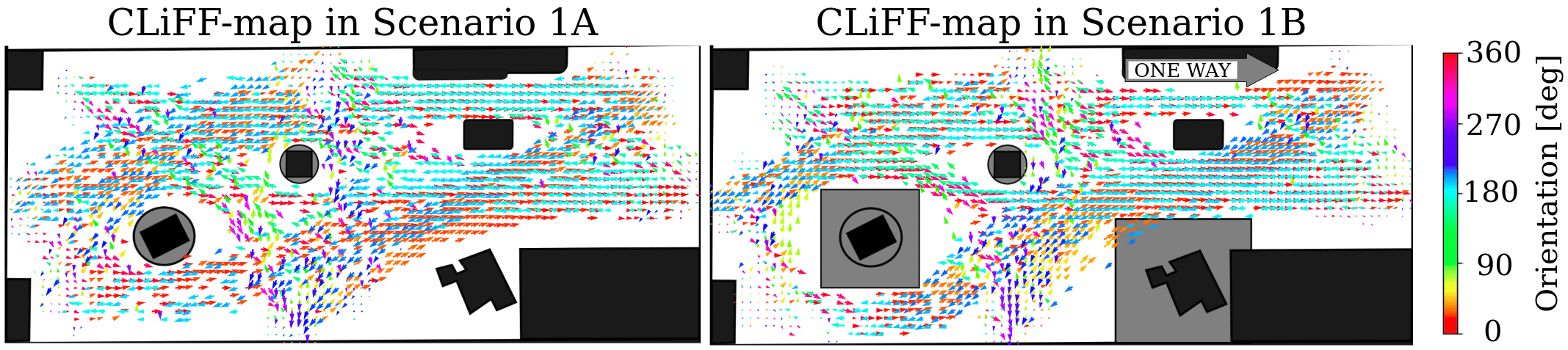}
    \caption{Maps of dynamics created from one day of data acquisitions. \textbf{Left:} Scenario~1A, as a baseline for human motion without semantic cues being present. \textbf{Right:} Scenario~1B layout, in which Gray areas around two of the robotic agents represent the lane markings to signalize areas of caution. A CLiFF-map \citep{kucner2020probabilistic} is used to represent statistical information about the flow patterns of humans in these settings.}
    \label{fig:SC1MoDs}
\end{figure}

Scenario~1 comprises two conditions: \textbf{condition A} involves static obstacles such as tables, stationary robots, and goal points. \textbf{condition B} introduces floor markings and stop signs in a one-way corridor in addition to the elements presented in \textbf{condition A}. The recording of \textbf{condition B} was before \textbf{condition A} to avoid biasing the participants towards the floor markings and to capture their natural reaction. Baseline \textbf{condition A} provides a clean environment without any floor markings or stop signs, allowing for the study of participants' motion patterns independently of these factors. This condition provides a foundation for understanding the effects of additional variables introduced in our other scenarios. Conditions A and B together enable the exploration of the impact of environmental cues on human motion (see Figure~\ref{fig:SC1MoDs}).


\subsubsection{Scenario 2: Role-Specific Motion Patterns in Industrial Environments}\label{scenario2}

Scenario~2 features the same environment layout as Scenario~1A (Figure~\ref{fig:SC1MoDs} left). In addition to the goal-driven navigation (\textbf{Visitors} role), this scenario introduces people performing different tasks as \textbf{Carriers}. For each run, we assign new roles to the participants. One participant carries small objects (i.e., buckets), and another carries medium objects (i.e., boxes) between two goal points. Finally, two participants move a large object (i.e., a poster stand). We use Discord\footnote{Free and easy-to-use communication and collaboration platform~\url{https://discord.com}} to instruct one member of the two-person team responsible for moving the large object. The usage of Discord enabled the dynamic allocation of new goal points and facilitated the coordination of participants' movements in this industrial context.  

In summary, this scenario presents role-specific tasks for participants and goal-driven navigation, creating a platform to study the impact of human occupation on their motion profiles and those of the other agents in a shared environment.

\subsubsection{Scenario 3: Impact of Mobile Robot Motion on Human Behavior}\label{scenario3}

\begin{figure}[!t]
\centering
\includegraphics[clip,trim=250mm 50mm 300mm 700mm, width=.6\linewidth]{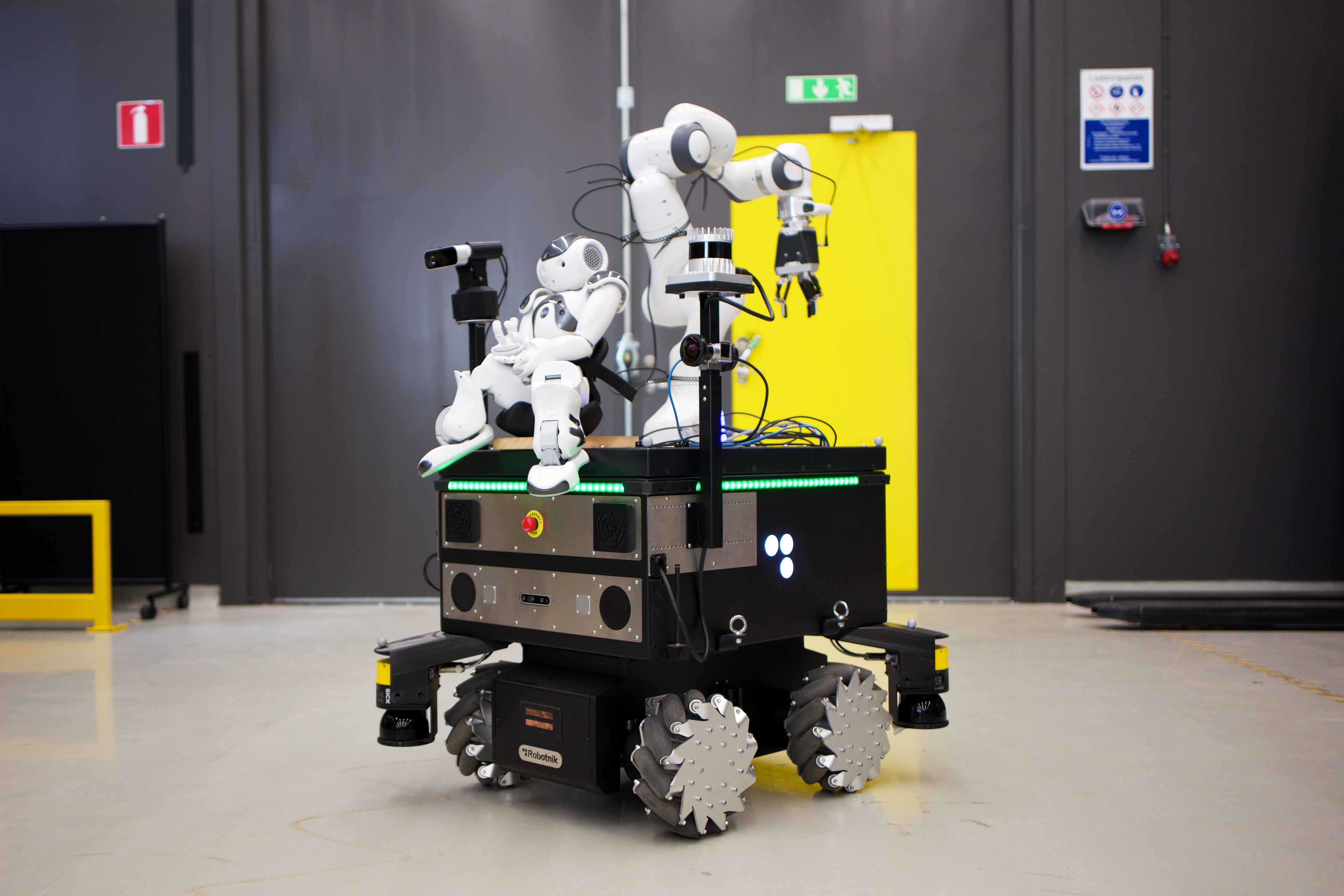}\\[2mm]\includegraphics[width=0.45\linewidth,height=4cm,keepaspectratio]{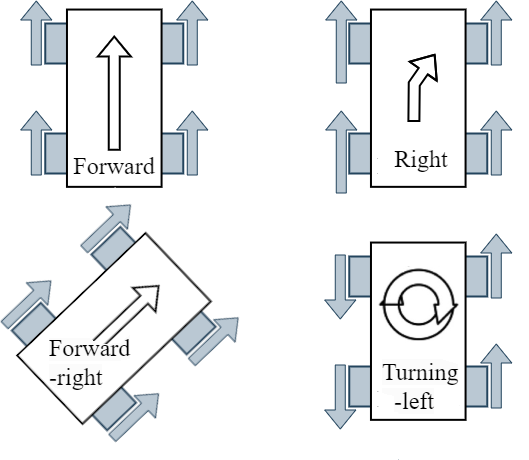}\hspace{0.5cm}
\includegraphics[width=0.45\linewidth,height=4.2cm,keepaspectratio]{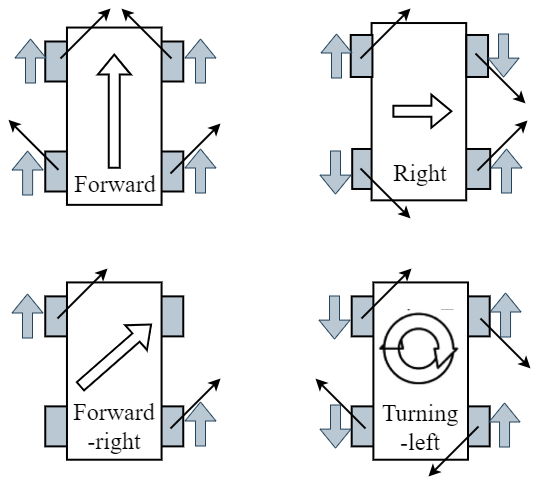}
\caption{Two types of mobile robot motion achievable with Mecanum wheels (\textbf{top}), the impact of these types on human behavior is explored in Scenario~3. \textbf{Left}: Differential driving where the two wheels on each side are synchronised to generate forward, backward, and turning motions. The grey arrows indicate the directions in which the individual wheel axis propel and the length of the arrow is proportional to the turning speed of the wheel. \textbf{Right}: Omnidirectional driving allows movement in any direction, including sideways and diagonally. Grey arrows indicate forward or backward propulsion of the individual wheels. The small black arrows indicate the normal vector of the direction in which the wheel pushes the robot. For more details, we refer the reader to \citet{tian2017research}.}
\label{fig:sidebyside}
\end{figure}

With Scenario~3, we introduce an opportunity to study the interplay between human activities and a mobile robot. In this scenario, the stationary DARKO robot of Scenarios~1 and 2 becomes mobile, exploring changes in the humans' motion patterns based on the mobile robot driving style. This scenario comprises two conditions, in which we modulated the way the mobile robot navigates: \textbf{condition A}, where the robot's motion always has a designated direction using directional differential-drive kinematics (see Figure~\ref{fig:sidebyside} bottom left) and \textbf{condition B}, where it can drive in any direction, i.e. omnidirectional (see Figure~\ref{fig:sidebyside} bottom right) using it's mecanum wheels (see Figure~\ref{fig:sidebyside} top). In both conditions, the roles of the participants remain the same as in Scenario~2, and a human operator controls the mobile robot using a remote controller to ensure the safety of the participants. Besides allowing for the study of human activities in the presence of a mobile robot, this setup also provides insights into how varying robot motion styles impact human behavior. 

\subsubsection{Scenario 4: Spatial HRI, Navigation in a Shared Environment}\label{scenario4}

This scenario includes participants with the roles of {\em Visitors--Alone HRI} and {\em Visitors--Group 2}, who freely move around a shared environment alongside the DARKO robot that navigates semi-autonomously. The robot moved autonomously, with the restriction of being supervised by an experimenter who could intervene and halt its movements via a controller.

Participants assigned with the role of {\em Visitors--Alone HRI} received instructions regarding a joint navigation task and engaged in interactions with the ARMoD. These participants take place in two conditions based on the interaction styles outlined in Section~\ref{robot_mode}, a Verbal-Only interactions style in \textbf{condition A} and a multi-modal one in \textbf{condition B}. Depending on the interaction style and the distance between goals one interaction lasted around 30--40 \si{\s}. If too many participants were at a goal point, the experimenter interrupted the mobile robot's autonomous navigation shortly before reaching the goal. If interrupted prematurely, the mobile robot told the participants to abort the interaction and continue drawinging cards. The mobile robot finished navigating autonomously to the goal point once it was less crowded.

Participants move either individually or in pairs between designated goal points. A specific card directs the individual participants ({\em Visitors-Alone HRI}) to approach the ARMoD and await further guidance. {\em Visitors--Group 2} are instructed to disregard this card. The experimenter controls ARMoD's behavior and sets the mobile robot's next goal point. Upon participants' arrival, ARMoD greets them and leads them jointly to the next goal point, where participants draw another card. 

\begin{figure}[!t]    
\includegraphics[width=0.6\linewidth,height=5cm,keepaspectratio]{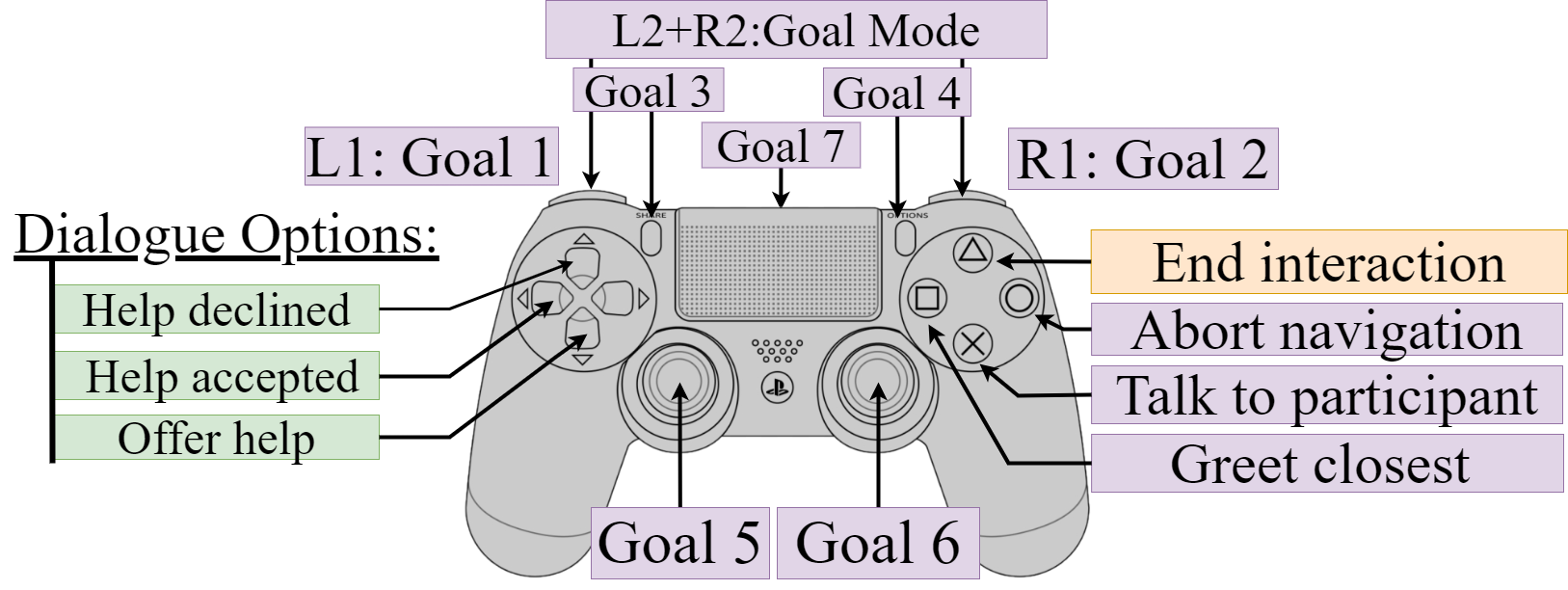}   \includegraphics[width=0.3\linewidth,height=5cm,keepaspectratio]{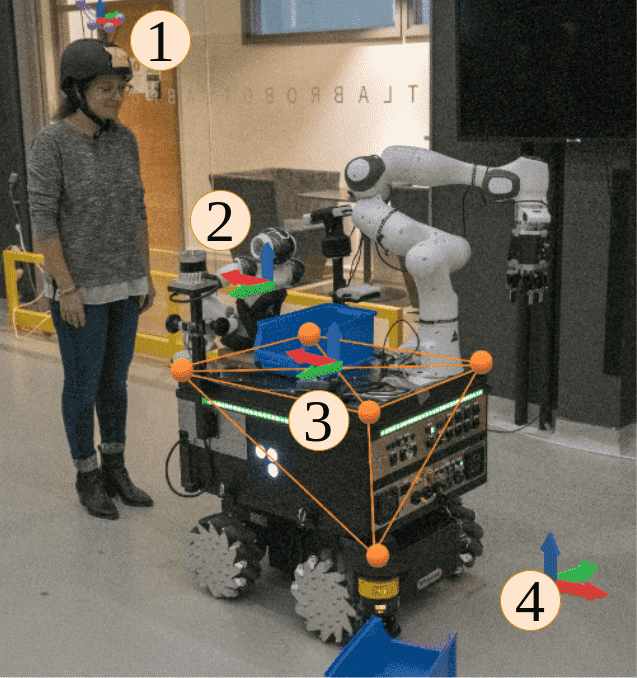}
\caption{\textbf{Left:} Input mapping used to control the ARMoD during HRI Scenarios~4 and 5. Purple items are used in both scenarios, yellow ones only in Scenario~4 and green ones only in Scenario~5. \textbf{Right:} Sample interaction with superimposed coordinate systems of: (1) Participant's Helmet,  (2) ARMoD-, (3) DARKO-, and (4) QTM-World Frame.}
\label{fig:HRIScenario}
\end{figure}

To ensure safe and seamless interactions, ARMoD's behaviors are triggered by an experimenter using a controller (see Figure~\ref{fig:HRIScenario} left). The experimenter initiates actions like ``Greet closest participant'' and ``Talking to participant'', guiding ARMoD's communication with participants. Concurrently, the mobile robot continues its autonomous navigation, albeit under the oversight of the experimenter, who can pause its movements if necessary. 

Accurate tracking of individuals was essential for facilitating seamless interactions between ARMoD and participants. Determining the ARMoD's position relative to individuals at any given moment, we leveraged the motion capture system's data, broadcasted into the local network using ``Robot Operating System (ROS)''  \citep{quigley2009ros}. This integration ensured precise transformations and provided position and orientation information, enabling ARMoD to accurately point, look, and establish eye contact with its interaction partners. Figure~\ref{fig:HRIScenario} right illustrates an interaction between a participant and ARMoD in this scenario. The position and orientation data of participants, robots, and the world frame are broadcasted within the local network, providing essential information to the path planner for DARKO and the interaction scheduler for ARMoD. In this figure, examples of established coordinate frames include (1) that of the helmets of participants defined based on the orientation of the marker, (2) a static coordinate frame for the ARMoD derived from the DARKO robots frame through an offset, (3) DARKO's coordinate frame, and (4) the motion capture reference's frame called the ``QTM-World Frame''.

This scenario investigates free movement in a shared environment alongside the DARKO robot, exploring semi-autonomous navigation. Participants engaged in interactions with ARMoD under varied conditions. These allow for a study of human-robot interactions, navigation tasks, and the impact of different interaction styles on participants' activities and movements.

\subsubsection{Scenario 5: Spatial Human Robot Interaction,  Proactive Robotic Assistance}\label{scenario5}
This scenario involves the roles: {\em Visitors--Alone}, {\em Visitors--Group 2}, and {\em Carrier--Storage Bin HRI}. The first two navigate between goal points by drawing cards. The {\em Carrier--Storage Bin HRI} takes on the role of a factory worker responsible for transporting storage bins, and interacting with DARKO through ARMoD. The experimenter controled ARMoD's behavior and supervised DARKO's motion for safety. During the interaction, ARMoD proactively offered assistance to the {\em Carrier--Storage Bin HRI}, informing them of the option to place a small storage bin on the mobile robot. If participants accepted, they could place the small storage bin on the DARKO robot for transportation between two designated points. The procedure described in Section~\ref{scenario4} enabled reliable perception of both human and robot positions for this scenario. This scenario features proactive assistance from a mobile robot to a human worker in a simulated factory environment.

\subsection{Participants Background and Priming}\label{subsec:prime}

\begin{figure}[!t]
    \centering
    \includegraphics[width=0.8\linewidth]{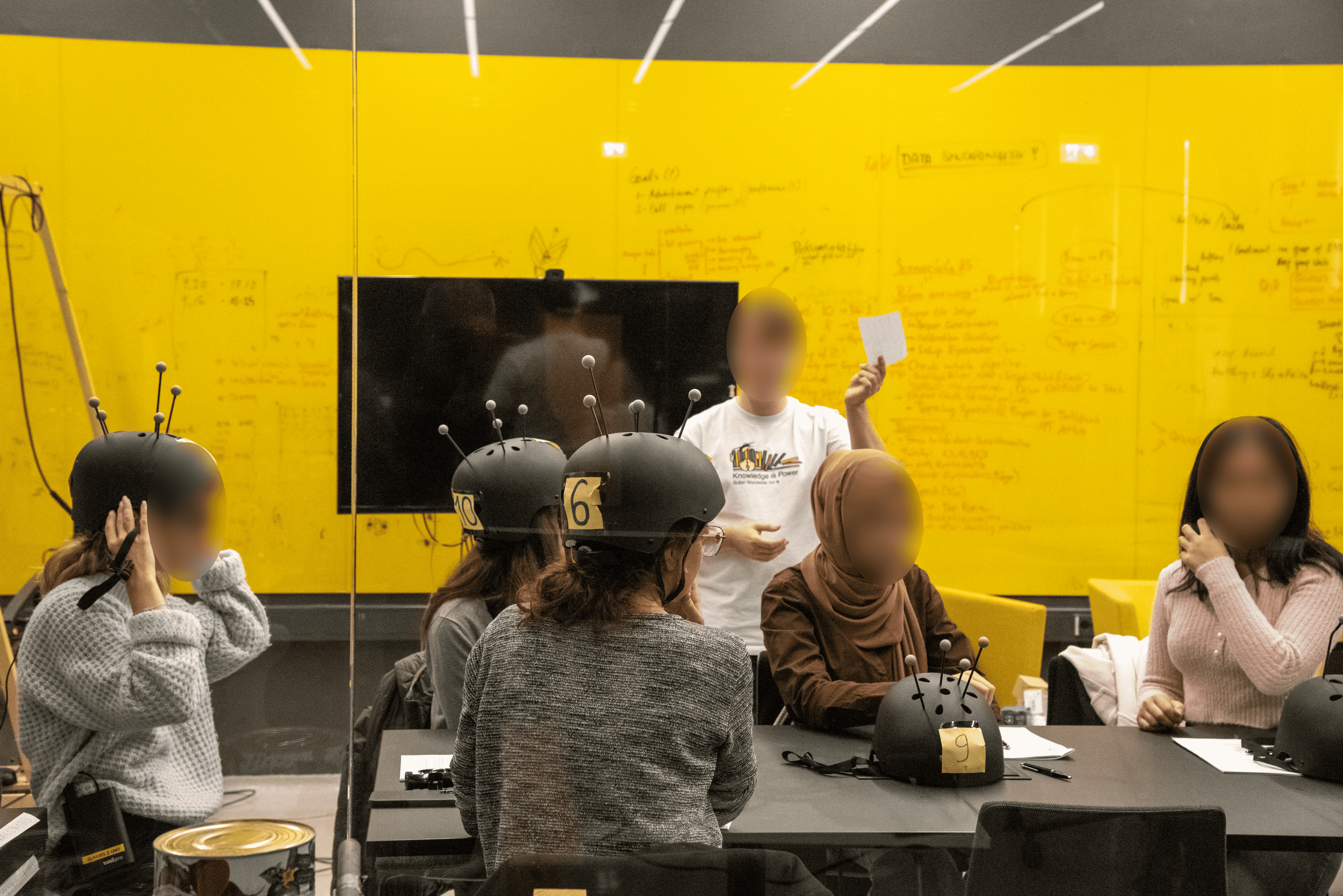}
    \caption{Initial priming of participants performed at the beginning of each recording day. Participants were instructed about the experimental setting and the recording procedure, including a briefing on the tasks, establishing familiarity with the equipment, as well as filling consent forms and an initial set of questionnaires.}
    \label{fig:optimus-prime}
\end{figure}

The average age of the participants was 30.18 years, with a standard deviation of 6.73, indicating a relatively homogeneous age group. The dataset contains a balanced gender distribution with 40 participants, of which 21 are male and 19 female. Geographically, 23 participants are from Sweden. From other European countries, there are 10, including the Czech Republic, Spain, Germany, and Italy, reflecting a diverse European representation. The remaining 7 participants come from countries on other continents like Asia, Africa, and South America, providing a broader international scope. We recruited the participants from different areas of the campus. Their backgrounds varied considerably, including differences in their highest academic degree and primary subjects. At the beginning of each recording day, participants completed a demographic questionnaire. We used this information to create diverse group compositions, aiming for optimal allocation of eye tracking devices across different roles (see Figure~\ref{fig:optimus-prime}). For example, we ensured that groups of two or three participants contained only one participant equipped with an eye tracker and the equipment of at least one of the carriers with an eye tracker.

During the data collection procedure, we guided the participants through a series of runs with specific instructions tailored to each scenario. Between successive runs, participants complete questionnaires while logistical preparations are made, such as removing floor markings, configuring a phone for voice chat using Discord (before Scenarios~2 and 3), monitoring and, if necessary, changing the batteries of eye trackers, and preparing the robots for Scenarios~3--5. After completing the questionnaire, participants are assigned new roles in Scenarios~2 and 3. We gave each group a new starting point for the next run, from which they drew their first card. Participants unfamiliar with their roles got a brief recap of their task-related responsibilities. In Scenario~3, we informed participants that an experimenter monitored the robot's motion for safety and teleoperated the robot. In Scenarios~4 and 5, participants were first briefed about their roles in the scenario (see Section~\ref{subsec:scenarios}) and then introduced to the ARMoD and the DARKO robot as co-workers in the room, with the ARMoD acting as a communicator on behalf of the DARKO robot.

After each run, participants completed the raw version of the NASA Task Load Index (RTLX)~\citep{Hart1988, Hart2006}. The scale consisted of a 21-point set of subscales [1 = low; 21 = high], each of which assessed the mental demand, physical demand, temporal demand, and frustration produced by the task as reported by the participant, as well as their self-perceived performance and frustration. After each session of the last run of Scenarios~3 or 5, participants complete two additional mobile robot questionnaires. First, they complete the Godspeed Questionnaire Series \citep{Bartneck2009}, a semantic differential set of subscales [5-point] that measures participants' perceptions of the robot in terms of anthropomorphism, animacy, likeability, perceived intelligence, and perceived safety. Second, they complete a 5-point Likert scale [1 = strongly disagree; 5 = strongly agree] to assess trust in the robot in industrial human-robot collaborations~\citep{charalambous2016}. Participants complete all questionnaires on paper.

\subsection{System Setup}\label{subsec:system}

\subsubsection{Hardware and Software Configuration}\label{subsec:config}

We used a motion capture system from Qualisys with ten infrared cameras (Oqus 7+) positioned around the room to track moving agents. The system provided comprehensive coverage of the room volume. Reflective markers arranged in distinct patterns of six degrees of freedom (6DoF) on bicycle helmets. These were tracked at \SI{100}{\Hz} with a spatial resolution of \SI{1}{\mm}. The coordinate frame of the system had its origin at ground level in the center of the room. Each participant and the robot are represented as unique rigid bodies (identifiable through the group of passive reflective markers arranged in specific patterns) in the system. This configuration enabled the precise capture of the 6DoF head position and orientation for each participant. We provided the participants with individualized helmets for the recording sessions. The specific helmet IDs used during each recording session are listed in Tables~\ref{tab:scenario1}, \ref{tab:scenario3}, and \ref{tab:scenario5} in the Appendix.

We captured eye tracking data using three distinct models of eye tracking devices: Tobii Pro Glasses~2 and 3, and Pupil Invisible. The Tobii Glasses models record raw gaze data at a frequency of \SI{50}{\Hz} and camera footage at \SI{25}{\Hz}, while the Pupil Glasses record gaze data at \SI{100}{\Hz} and camera footage at \SI{30}{\Hz}. To export Tobii Glasses data, we used the I-VT Attention filter, which is optimized for dynamic situations, to classify gaze points into fixations and saccades based on a velocity threshold of $100\degree/s$.  All eye trackers are equipped with an IMU that comprises an accelerometer and a gyroscope operating at \SI{100}{\Hz}. In addition, the Tobii Glasses 3 are equipped with a magnetometer that operates at \SI{10}{\Hz}. The infrared cameras in these devices capture the human gaze, which is then superimposed onto a 2D video by the scene cameras. The Pupil Invisible Glasses' scene camera has a resolution of $1088\times1080$ pixels, with both horizontal and vertical field of view (FOV) angles measuring $80\degree$. In contrast, the Tobii Glasses offer a resolution of $1920\times1080$ pixels. The Tobii~3 Glasses feature FOV angles of $95\degree$ horizontally and $63\degree$ vertically, while the FOV of the Tobii~2 Glasses $82\degree$ horizontally and $52\degree$ vertically.

The DARKO robot integrates several sensors, including an Ouster OS0-128 lidar, two Azure Kinect RGB-D cameras (one of which was used in these recordings), two Basler fish-eye RGB cameras, and two Sick MicroScan 2D safety lidars. The Azure Kinect cameras have a resolution of $2048\times1536$ at \SI{6}{\Hz}, a horizontal field of view of $75\degree$, and a tracking range of up to \SI{5}{\m}. The Basler fish-eye RGB cameras have a resolution of $1700\times1536$ at \SI{20}{\Hz}.
The DARKO robot is augmented with a NAO robot acting as ARMoD for participant interaction. The NAO is attached to a seat on the DARKO robot, facilitating the communication of spatial motion intent. This arrangement aligns the ARMoD's body orientation with the direction of movement in scenarios where DARKO employs a directional driving style.

Recordings from the DARKO robot and the motion capture system were synchronized using ROS timestamps. Taking advantage of the integration of the motion capture system with ROS 1 Melodic, we recorded all of the robot's on-board sensor data and the 6DoF positions of the people using ROS bag files and in text form. 

\subsubsection{Sensor Calibration}\label{subsec:calibrate}

The precision of the data acquisition relied on sensor calibration procedures to ensure accurate measurements and reliable data interpretation throughout the experiments. In this section, we provide a detailed description of our calibration methods for both the motion capture system and the eye tracking devices. We followed separate calibration routines for each sensor. These calibration routines allowed for the robustness and reliability of our dataset, allowing for accurate analysis and interpretation of participants' behaviors and interactions within the recorded scenarios.

\begin{figure}[!t]
    \includegraphics[width=0.48\linewidth,height=4cm,keepaspectratio]{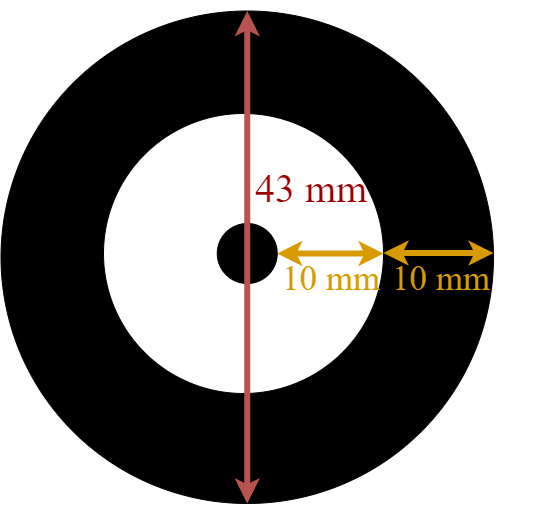}
    \includegraphics[width=0.48\linewidth,height=5cm,keepaspectratio]{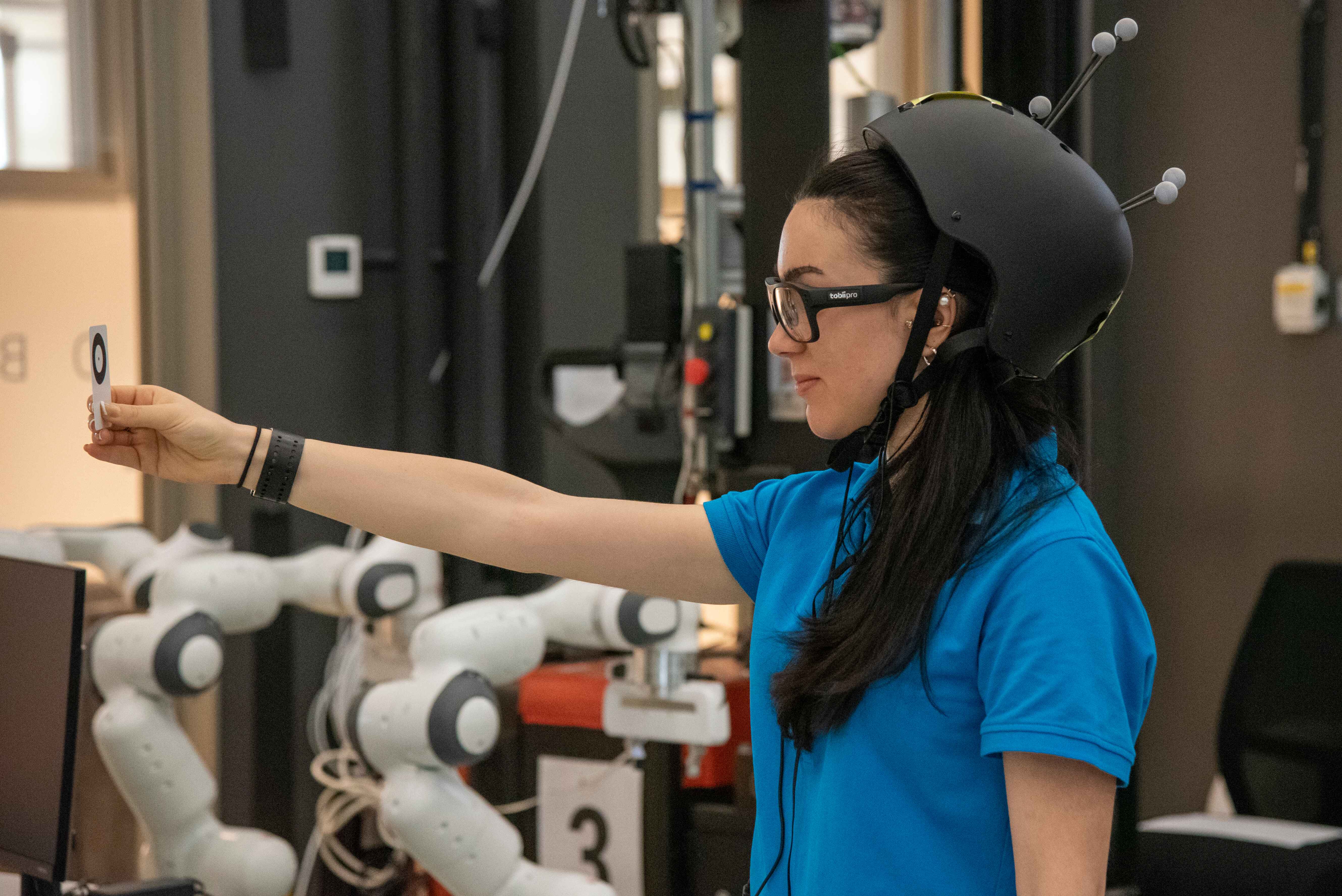}
    \caption{\textbf{Left:} Calibration pattern encompassing circles of different sizes printed on the card used for calibrations. Outer circle has a diameter of \SI{43}{\mm}. The radii of the inner the two circles \SI{20}{\mm} and \SI{3}{\mm} for the smallest center circle \textbf{Right:} A calibration procedure for mobile eye tracking glasses. The participant stands and holds a card with a black dot at eye level, about an arm’s length away. The participant focuses on the dot to align the eye tracking system with their eye movements. This step is essential to account for individual differences in eye anatomy and behavior.}
    \label{fig:tobii_calibration}
\end{figure}

For the eye tracking devices we followed the calibration procedures for both Tobii Glasses models (see Figure~\ref{fig:tobii_calibration}) as outlined in their respective user manuals to optimize eye tracking accuracy (see \cite{TobiiGlasses2Manual} and \cite{TobiiGlasses3Manual}). This process involved positioning a calibration target, ensuring its visibility, having participants focus on its center. To ensure accurate recordings with the Pupil
Invisible Glasses, we followed  best calibration practices outlined by- and validated the calibrations with the dedicated software of \cite{Core2023}.

To ensure data accuracy of the motion capture system, rigorous daily calibration routines were performed prior to the start of each recording session. We used the standard calibration kit with a \SI{502.2}{\mm} carbon fibre wand to fine-tune the system. These calibrations allowed us to define precise rigid bodies that enabled 6DoF tracking. This approach ensured the accurate capture of spatial dimensions (X, Y, Z) and rotational elements (roll, pitch, yaw) of objects within the 3D environment, resulting in an average residual tracking error of \SI{2}{\mm}. Rigid bodies of helmets and objects such as the large object for the cariers or the DARKO robot, were strategically designed to enable simultaneous and highly accurate capture of all object poses and locations. 

\subsection{Post Processing}\label{subsec:align}

Multi-modal data synchronization was necessary in our data collection. We used ROS and custom Python scripts to align the data streams while maintaining temporal integrity. To achieve synchronicity between the motion capture and eye tracking data, we strategically placed custom events associated with precise timestamps in the two data streams using the respective software of the eye tracking devices such as Tobii Pro Lab \citep{TobiiProLabManual} and Pupil Player \citep{PupilLabs2023} as well as the Qualisys Track Manager (QTM) \citep{QTMUserManual} for the motion capture system. This procedure resulted in CSV files where all modalities' timestamps are synchronized on the motion capture system's timestamp. Within these files, eye tracking data is available for frames where all markers of the rigid body are tracked by the motion capture system, as it is a prerequisite to determine the 3D gaze vector using a correct head orientation. The frame numbers for each respective eye tracker's scene recording are indexed in the column named ``SceneFNr'' in the corresponding CSV file.

To facilitate a thorough analysis of the eye tracking data in our study, we offer access to the raw data from the Tobii glasses, along with essential synchronization details. To ensure data protection, the scene recordings are provided in a blurred format and removed audio data. Access to the raw data from the Pupil Invisible glasses can be granted upon individual request, ensuring careful and ethical distribution of sensitive data.

An extensive post-processing stage, including the above mentioned synchronization and alignment, followed the data acquisitions. Its aim was to refine and validate the collected data as well as to ensure the protection of sensitive data. This stage involved several key procedures, such as eliminating artifacts and noise caused by marker occlusion, lighting variations, and camera disruptions. We also rectified misidentified trajectories through a combination of spatial and temporal consistency evaluations, with manual adjustments applied when needed.
\section{Working with the TH\"{O}R-MAGNI Dataset}\label{sec:usage}
\subsection{Data Formats}\label{subsec:format}

For the purpose of dissemination, the dataset has been categorized into five recording scenarios (see Section~\ref{sec:dataset} for a detailed description), aligning with the respective days of data collection. Each scenario's data is organized into separate folders. Within each folder, multiple acquisitions conducted over the five days of recording are stored. The folders corresponding to the first three scenarios (1--3) contain acquisitions from four days (in May 2022), while the folders representing the last two scenarios (4 and 5) encompass recordings from one day (in September 2022). To enhance the diversity of motion data in the recordings and mitigate random artifacts, we record multiple runs for each scenario and condition. It is essential to note that all files are intended to be extracted into a common directory. In this way, the arrangement preserves the temporal structure of the recorded data. 

Each run's data includes a CSV file and up to two .mp4 videos representing the recordings from the scene cameras of the Tobii eye trackers and if the robot was in motion during the Scenarios 3--5 continuous 3D point clouds from the Ouster lidar as well as the RGB videos from one of the fish-eye cameras. The structure of the recorded data is shown in the Tables~\ref{tab:scenario1}, \ref{tab:scenario3}, and {\ref{tab:scenario5}} in the Appendix. In the following subsections we will provide more specific details on the usage and processing of the individual files. 

\subsubsection{Comma-Separated Value Files}
Each CSV file contains a header with important metadata, including number of frames for the recording, rigid body and marker details, units of measurement, role labels, and eye tracking specifics (see Table~\ref{tab:CSVHead} in the Appendix). The rest of the CSV files contains the merged data from the motion capture system and the eye tracking devices, organized based on the rigid bodies of the participants' helmets. Thus, the data of each rigid body is organized into columns containing the XYZ coordinates of all markers (e.g., ``Helmet\_1 – 2 X'' indicating the data for helmet one, marker two and axis X), XYZ coordinates of the centroid of all markers, 6DOF orientation of the rigid body's local coordinate frame, and, if available, eye tracking data including 2D gaze coordinates, 3D gaze vectors, frame number of scene recording, eye movement types (such as saccades or fixations), and IMU data (accelerometer, gyroscope, and magnetometer). 

Missing data is indicated by either ``N/A'' (not available) or an empty cell. The temporal indexing in these files is provided by the ``Time'' or ``Frame'' column, which indicates the timestamp or frame number of the motion capture system, respectively. 

\subsubsection{Robot sensor data}
The sensor data from the robot includes lidar data and videos captured by the Azure Kinect camera and the Basler camera. Lidar 3D point clouds are provided in the Point Cloud Data (PCD) file format, corresponding to each timestamp. 
The lidar data for each run is supplied in a zip file, which is labeled with the same File ID as referenced in the Tables~\ref{tab:scenario1}, \ref{tab:scenario3}, and \ref{tab:scenario5}  in the Appendix.
In terms of video data, both the RGB-D and fish-eye camera video streams are unrectified, providing raw visual data, and are only available upon request to ensure a suitable data protection.

\subsubsection{Additional data}

In addition to the CSV files containing information about the recorded data from the eye trackers and the motion capture system, we provide the scene recordings from most of the Tobii eye tracking devices as .mp4 videos.  The videos of the scene recordings were carefully post processed, as we blurred all faces of the participants using a dedicated video-redaction software (``Caseguard'') to ensure data protection. The raw camera video from the Pupil Invisible Glasses scene has distortions that need to be corrected. For this purpose, we provide JSON files with the necessary intrinsic camera parameters to compensate. All data from the Pupil Invisible eye tracking devices and the remaining data from the Tobii devices are available upon request.

\subsection{Development Tools}\label{subsec:tools}

The majority of existing datasets in the field lack a dedicated toolbox for streamlined visualization and preprocessing. Addressing this gap, we contribute a set of data visualization tools, including a dashboard, and introduce a specialized Python package named {\em thor-magni-tools}. This package is designed to facilitate the filtering and preprocessing of raw trajectory data, enhancing the accessibility and usability of the TH\"{O}R-MAGNI dataset. By making available these resources, we aim to provide researchers with versatile and fast means to navigate, analyze, and extract valuable insights from the dataset.

\subsubsection{Data Visualization} In order to provide researchers and users with an intuitive interface for the exploration of human movement, gaze patterns, and environmental perception of the TH\"{O}R-MAGNI dataset, we made a set of visualization tools publicly available\footnote{\url{https://github.com/tmralmeida/magni-dash/tree/dash-public}}. 
Our visualization dashboard provides a user-friendly interface with multiple interactive components. The dashboard includes the following key features:
\begin{enumerate}
    \item \textbf{Trajectory visualization}: users can visualize agents' trajectories in 2D or 3D space. 
To represent different agents, the trajectories are color-coded, which allows the user to identify patterns and variations.
    \item \textbf{Velocity profiles}: the dashboard displays also velocity profiles corresponding to each trajectory, allowing users to analyze 
    speed variations during different phases of movement. This feature helps to understand 
the dynamics of human movement under different conditions.
    \item \textbf{Eye tracking data alignment}: gaze data is overlaid on the 3D trajectories. This provides insight into visual attention during different phases of motion. Researchers can explore how gaze patterns align with specific trajectory segments, promoting the study of the cognitive processes underlying human actions.
    \item \textbf{Lidar data visualization}: lidar sensor data is presented in 3D format to show the environmental context of human motion. 
    This information is critical for studying lidar-based human detectors onboard mobile robots, especially in complex environments like in TH\"{O}R-MAGNI.
\end{enumerate}

In addition to data visualization, our dashboard contains concise scenario descriptions. 
Each scenario represents a unique context in which human motion data was captured (described in  Section~\ref{subsec:scenarios}). 
These descriptions include information such as the physical environment, task objectives, social interactions, and any 
specific conditions imposed on the participants (e.g., transporting objects between two goal points). Understanding these scenarios is vital for the accurate interpretation of the data and ensures that researchers can contextualize their analyses effectively.

\subsubsection{Data Filtering and Preprocessing with {\em thor-magni-tools}}

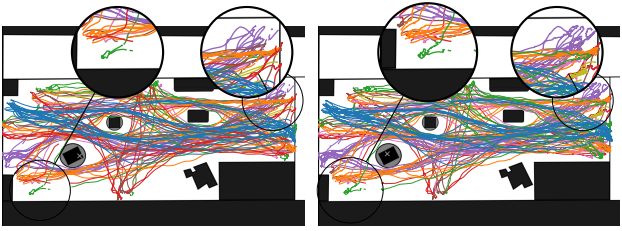
\begin{figure}[!t]
    \centering
    \begin{tikzpicture}
        [spy using outlines={
              shape=chamfered rectangle,
              every spy on node/.append style={inner sep=+1pt},
              every spy in node/.append style={inner sep=+1pt},
              connect spies
            }]

    \node[inner sep=0, anchor= south west](image) at (0,0){\includegraphics[width=\linewidth]{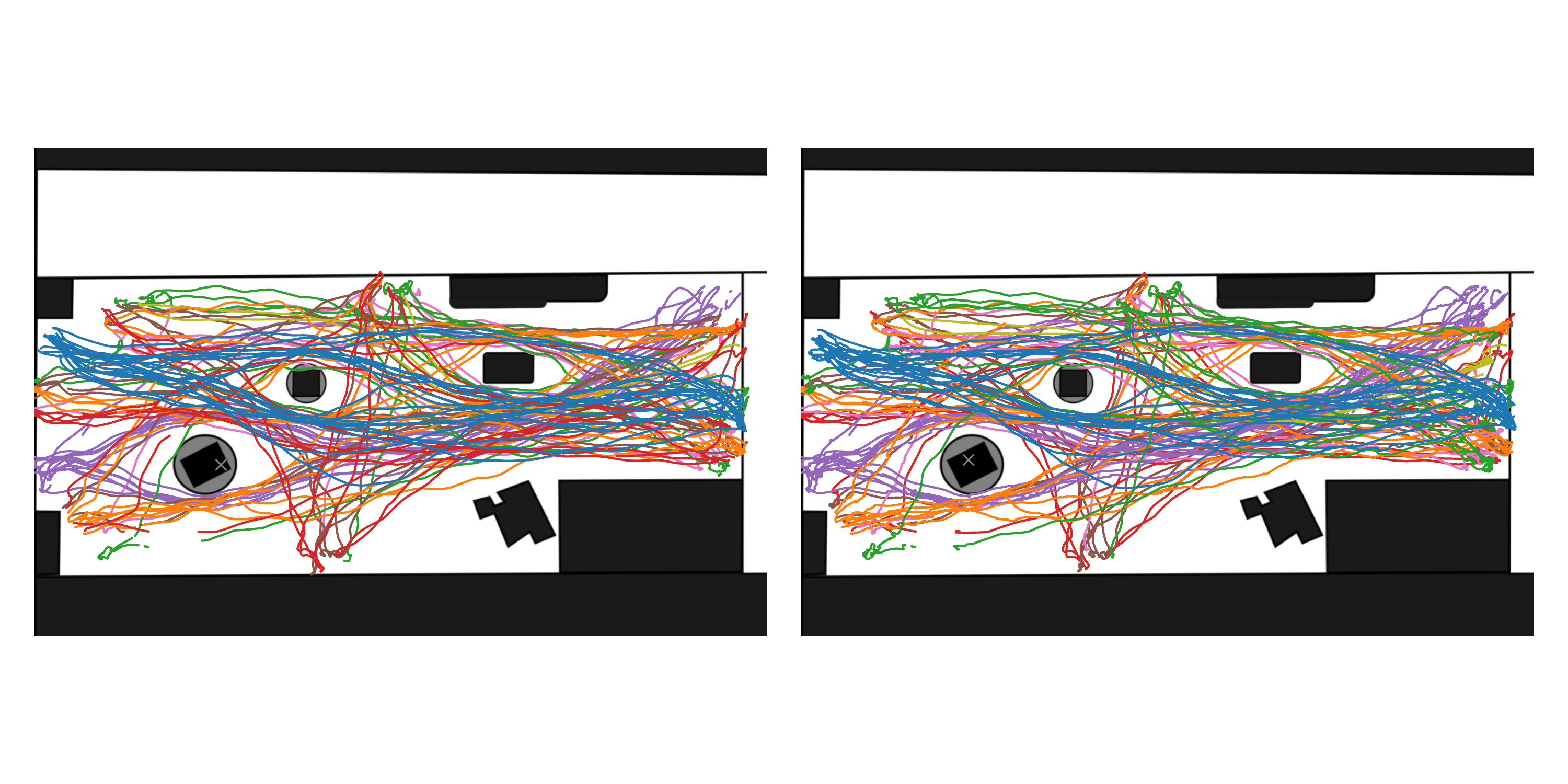}};
    \begin{scope}[x={(image.south east)},y={(image.north west)}]
        \coordinate (focushere1)  at (0.44,0.58);  
        \coordinate (zoom1)       at (0.4,0.73);   
        \spy[circle, width = 1.2cm,black,magnification=1.5] on (focushere1) in node at (zoom1);

        \coordinate (focushere2)  at (0.08,0.30);  
        \coordinate (zoom2)       at (0.2,0.73);   
        \spy[circle, width = 1.2cm,black,magnification=1.5] on (focushere2) in node at (zoom2);

        \coordinate (focushere3)  at (0.56,0.30);  
        \coordinate (zoom3)       at (0.68,0.73);   
        \spy[circle, width = 1.3cm,black,magnification=1.5] on (focushere3) in node at (zoom3);

        \coordinate (focushere4)  at (0.92,0.58);  
        \coordinate (zoom4)       at (0.88,0.73);   
        \spy[circle, width = 1.2cm,black,magnification=1.5] on (focushere4) in node at (zoom4);

    \end{scope}
    \end{tikzpicture}
    \vspace{-15mm}
    \caption{Filtering methods in a 4-minute recording from Scenario 1. \textbf{Left:} trajectories filtered using the most reliable marker. \textbf{Right:} trajectories filtered using the average of the 
    tracking locations of each marker. 
    Although the average tracking markers method provides longer tracks, it induces jerkier trajectories, 
    especially near the boundaries of the motion capture volume (e.g. bottom left and top right).}
    \label{fig:filtering_types}
\end{figure}

To facilitate the use of the agents' trajectories in our dataset, we employed the 
{\em thor-magni-tools} Python package\footnote{\url{https://github.com/tmralmeida/thor-magni-tools}}, 
a tool designed specifically for filtering, preprocessing, and visualizing trajectory data. 
This tool focuses on mitigating tracking issues arising from the motion capture system, enhancing the quality of the data for downstream tasks such as studying novel trajectory prediction methods. To filter 3D trajectory data, we provide two methods: (1) using the most reliable marker, i.e., the marker of each helmet with the highest number of tracking locations and (2) restore the helmet tracking based on the average of the tracking locations of each marker. Both approaches offer a trade-off between tracking quantity and quality. The method utilizing the best marker exclusively produces smoother trajectories due to its reliance on a single marker. Conversely, the method averaging the positions of all visible markers generates longer trajectories but with increased jerkiness, as it incorporates data from multiple markers, which can vary. However, this jerkiness can be alleviated through the application of a moving average filter in subsequent processing stages. Figure~\ref{fig:filtering_types} shows an example of the two methods applied on TH\"{O}R-MAGNI trajectory data.

\begin{figure}[!t]
    \centering
    \begin{tikzpicture}
        [spy using outlines={
              shape=chamfered rectangle,
              every spy on node/.append style={inner sep=+1pt},
              every spy in node/.append style={inner sep=+1pt},
              connect spies
            }]
            \node[inner sep=0, anchor= south west](image) at (0,0){\includegraphics[width=\linewidth]{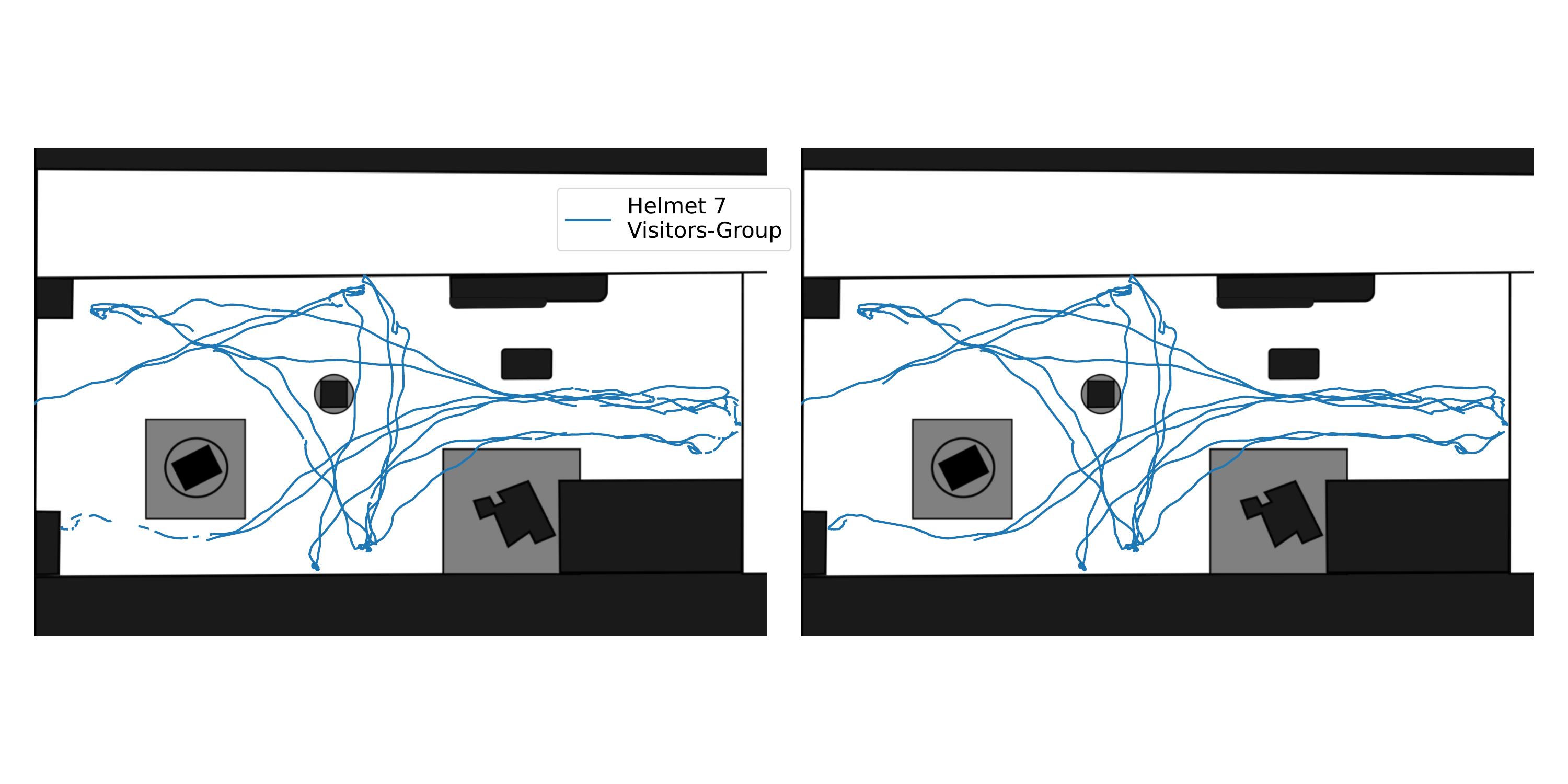}};
            \begin{scope}[x={(image.south east)},y={(image.north west)}]
                \coordinate (focushere1)  at (0.45,0.43);  
                \coordinate (zoom1)       at (0.3,0.73);   
                \spy[circle, width = 1.2cm,black,magnification=1.8] on (focushere1) in node at (zoom1);

                \coordinate (focushere2)  at (0.93,0.43);  
                \coordinate (zoom2)       at (0.88,0.73);   
                \spy[circle, width = 1.2cm,black,magnification=1.8] on (focushere2) in node at (zoom2);

                \coordinate (focushere3)  at (0.07,0.33);  
                \coordinate (zoom3)       at (0.1,0.73);   
                \spy[circle, width = 1.3cm,black,magnification=1.5] on (focushere3) in node at (zoom3);

                \coordinate (focushere4)  at (0.56,0.33);  
                \coordinate (zoom4)       at (0.65,0.73);   
                \spy[circle, width = 1.3cm,black,magnification=1.5] on (focushere4) in node at (zoom4);

            \end{scope}
    \end{tikzpicture}
    \vspace{-15mm}
    \caption{Example of a 4-minute Helmet trajectory in Scenario 1.
   \textbf{Left:} raw trajectory data depicting gaps, especially around extreme environmental locations. \textbf{Right:} post-processed tracing with 100 maximum positions without tracking (\SI{1}{s}) interpolation, 
    showcasing enhanced continuity and completeness in the trajectory.}
    \label{fig:thor_magni_tools}
\end{figure}
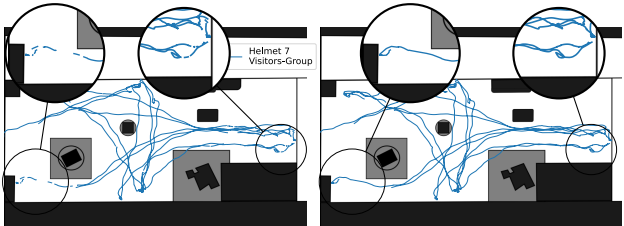

For both 3D and 6D tracks (X, Y, Z, and 3D orientation), we provide an interpolation method 
based on a predefined maximum number of positions in the absence of tracking. This method is used to fill in the missing data points while maintaining the integrity of the motion patterns and ensuring continuity in the trajectories. An example of the interpolation of a trajectory based on {\em thor-magni-tools} is depicted in Figure~\ref{fig:thor_magni_tools}. Finally, this tool also offers optional preprocessing steps, including downsampling and signal smoothing through a moving average filter, further refining the processed trajectories.

\section{Analysis and Comparison to Existing Human Motion Datasets}\label{sec:results}

This section presents a comparison with popular human trajectory datasets, specifically the ETH/UCY benchmark and TH\"{O}R, with our TH\"{O}R-MAGNI dataset. Our analysis encompasses a multidimensional evaluation, covering various facets of the data recordings. These include trajectory continuity, social proxemics delineating interpersonal interactions, as well as motion characteristics such as velocity profiles and trajectory linearity. Through this comparison, we aim to situate TH\"{O}R-MAGNI among its predecessors, showing its potential for advancing human motion analysis and human-robot interaction research. 

\subsection{Metrics for Trajectory Data Comparison}

To evaluate the trajectory data of our dataset in comparison to previous data collections,
we employ metrics proposed by \citet{thor_20} and \citet{amorian21}:

\begin{itemize}
    \item \textbf{Tracking Duration}~(\SI{}{s}): this metric represents the average duration of continuous tracking for 
    all human agents. A higher value indicates longer tracking, which is favorable 
    for long-term human motion prediction methods.
    \item \textbf{Minimal Distance Between People}~(\SI{}{m}): this metric measures the minimum distance 
    observed between individuals in the dataset. It provides insights into the proximity of 
    human agents during their interactions, offering valuable data for studies related to personal 
    space (proxemics) and social dynamics.
    \item \textbf{Number of 8-second Tracklets}: this metric counts the non-overlapping tracklets of 
    8-second duration after downsampling to \SI{0.4}{s} and applying a moving average filter. 
    These choices align with current trajectory prediction 
    benchmarks such as those outlined in \citet{kothari_21}. These tracklets offer discrete 
    temporal segments for analysis, ensuring compatibility with existing evaluation standards 
    in the field of trajectory prediction.
    \item \textbf{Motion Speed}~(\SI{}{m}/{}{s}): motion speed represents the velocity of all human agents. 
    A higher standard deviation in motion speed indicates a diverse range of behaviors within 
    the dataset. This diversity is essential for capturing various movement patterns and is 
    crucial for robustness in trajectory prediction models.
    This metric is computed in the 8-second tracklets.
    \item \textbf{Path Efficiency}: path efficiency quantifies the linearity of trajectories in the dataset, 
    ranging between 0 and 1 \citep{amorian21}. It is calculated by dividing the distance between the first and 
    last points by the cumulative distance traveled. A lower coefficient suggests more complex 
    and non-linear trajectories, providing valuable insights into intricate human movement patterns.
    This metric is computed in the 8-second tracklets.
\end{itemize}

\subsection{Trajectory Data Comparison}

We compare our dataset with the TH\"{O}R dataset and the ETH/UCY trajectory prediction benchmark.
The TH\"{O}R dataset encompasses three distinct scenarios, each featuring participants performing different tasks such as individual and group movement, as well as box transportation, different amounts of obstacles, and a mobile robot in the environment. In TH\"{O}R Scenario~1 (TH\"{O}R-S1), participants navigate the environment with one static obstacle. TH\"{O}R Scenario~2 (TH\"{O}R-S2) introduces a mobile robot navigating around the static obstacle, while participants continue their tasks. Finally, in TH\"{O}R Scenario~3 (TH\"{O}R-S3), the mobile robot becomes a static obstacle and an additional obstacle is added to the scene.
The ETH/UCY trajectory prediction benchmark consists of five scenes: ETH, HOTEL, UNIV, ZARA1, and ZARA2. These scenes represent five outdoor public spaces that capture natural human motion patterns, resulting in a benchmark widely used by the human trajectory prediction community~\citep{salzmann20,dendorfer21,yue22,almeida_pred23}.

\begin{figure}[!t]
\includegraphics*[width=\linewidth]{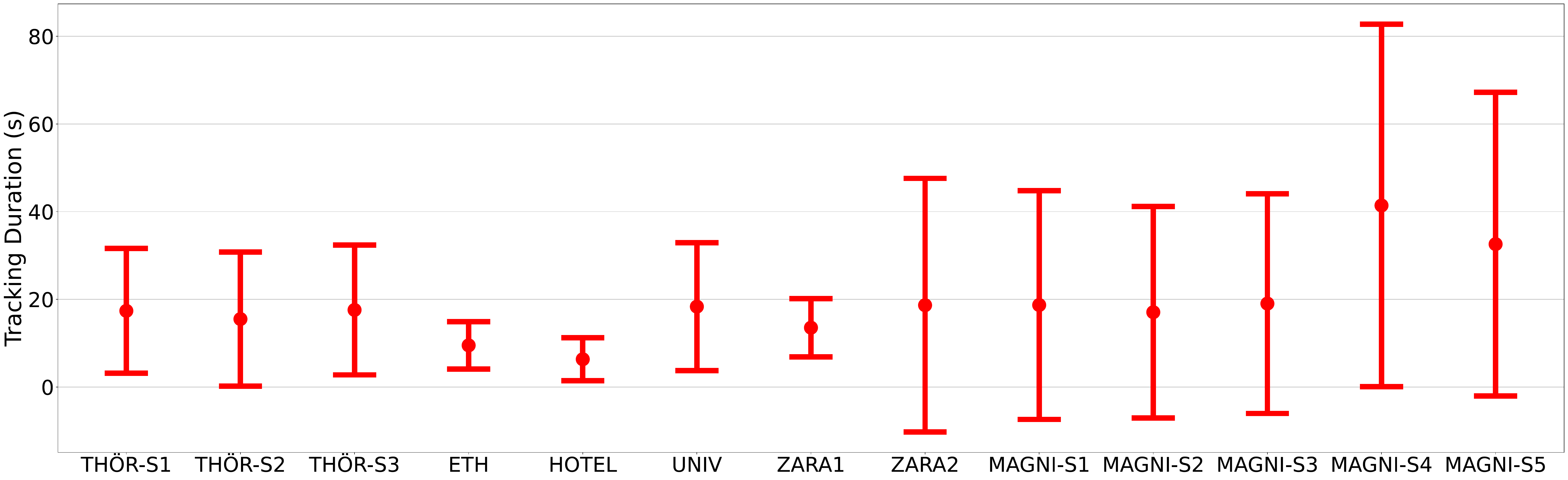}
\caption{Tracking durations (mean ± one standard deviation) across datasets in seconds. Scenarios~1--3 of TH\"{O}R-MAGNI provide comparable tracking durations to previous datasets, while Scenarios~4 and 5 provide longer tracks.}
\label{fig:track_durations}
\end{figure}

Firstly, we show the tracking durations in Figure~\ref{fig:track_durations}.
TH\"{O}R presents consistent average tracking durations around 15.5 to 17.6 seconds across the three scenarios. 
In contrast, TH\"{O}R-MAGNI shows wider variations, for instance, Scenario~4 features longer tracking durations 
(averaging 41.3 seconds), whereas Scenario~2 has the shortest durations (averaging 17.1 seconds).
This variability can be attributed to the density of participants; that Scenarios~4 and 5, involved fewer human agents in a smaller space, may contribute to higher quality tracking. Nevertheless, TH\"{O}R-MAGNI 
has comparable or higher tracking time than TH\"{O}R. Furthermore, compared to the ETH/UCY benchmark (i.e., ETH, HOTEL, UNIV, ZARA1, and ZARA2 scenes), TH\"{O}R-MAGNI offers comparable or significantly longer tracking durations. This makes our dataset more useful than its predecessors for tasks such as long-term human motion prediction and human-robot interactions.

\begin{figure}[!t]
\includegraphics*[width=\linewidth]{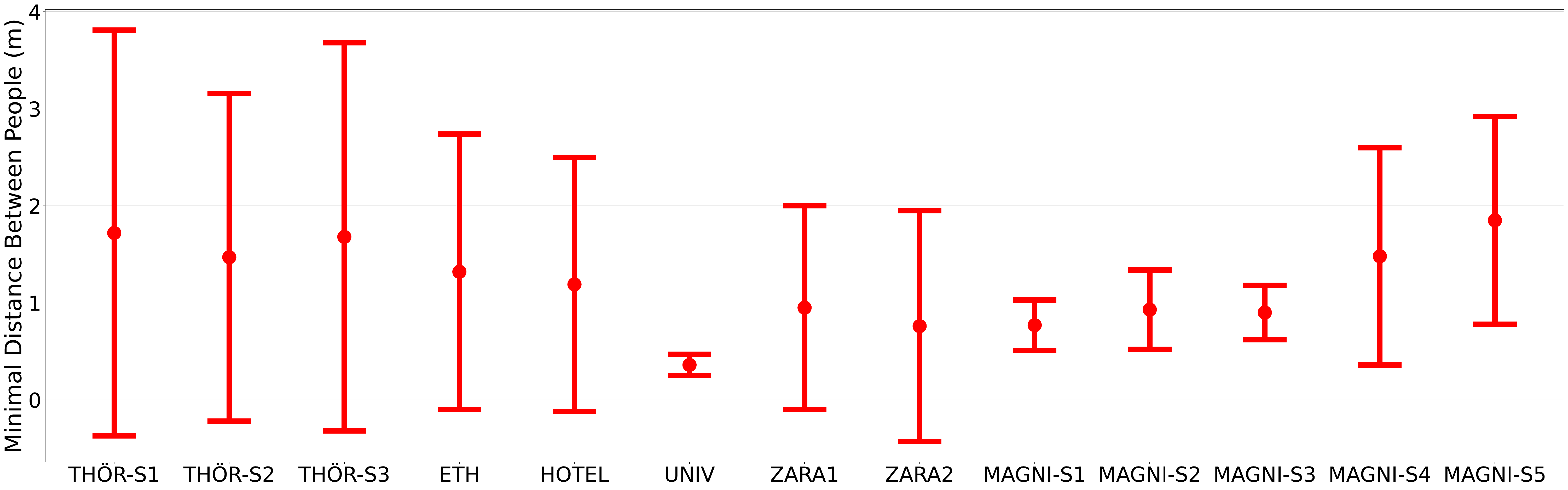}
\caption{Minimal distance between people (mean ± one standard deviation) across datasets in meters. Lower spatial navigational freedom in Scenarios~1--3 of TH\"{O}R-MAGNI potentiates reduced social distances between participants. These results are more consistent with the ZARA1 and ZARA2 scenes, while Scenario~4 and 5 (with more spatial freedom) show similar results to TH\"{O}R, ETH, and HOTEL datasets.}
\label{fig:min_dist}
\end{figure}

Secondly, we compare the minimal distance between people in Figure~\ref{fig:min_dist}. Again, human density plays an important role: TH\"{O}R-MAGNI Scenarios~1--3 show low values comparable to those in ZARA1/ZARA2, while Scenario~4 and 5 reach values similar to TH\"{O}R, ETH, and HOTEL. The higher participant density in TH\"{O}R-MAGNI Scenarios~1--3 results in reduced spatial navigational freedom, leading to increased interactions and decreased social 
distances between individuals. 

\begin{figure}[!t]
\includegraphics*[width=\linewidth]{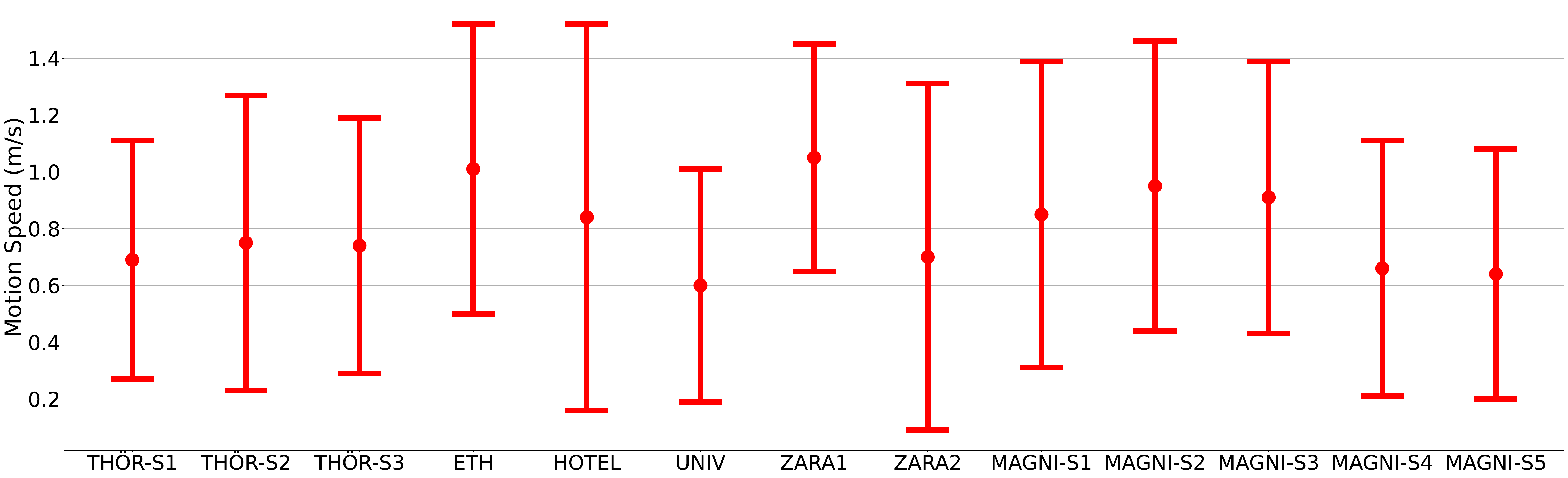}
\caption{Motion speed (mean ± one standard deviation) for 8-second tracklets across datasets in meters per second.}
\label{fig:speed}
\end{figure}

Thirdly, the motion speed statistics are shown in Figure~\ref{fig:speed}. Despite the higher participant density in Scenarios~1--3 of TH\"{O}R-MAGNI, these datasets feature 
faster human agent navigation than TH\"{O}R and akin to those in ETH, HOTEL, and ZARA1 scenes, possibly influenced by the task of object transportation, 
impacting their velocity profiles. Participants in Scenarios 4 and 5 of TH\"{O}R-MAGNI have an average velocity similar to those in TH\"{O}R, UNIV and ZARA2. Also, generally, TH\"{O}R-MAGNI shows comparable standard deviations in motion 
speeds, indicating diverse and varied movement patterns among human agents.
The similarity of the velocity profiles to previous datasets suggests that our dataset is also natural and diverse.

\begin{figure}[!t]
\includegraphics*[width=\linewidth]{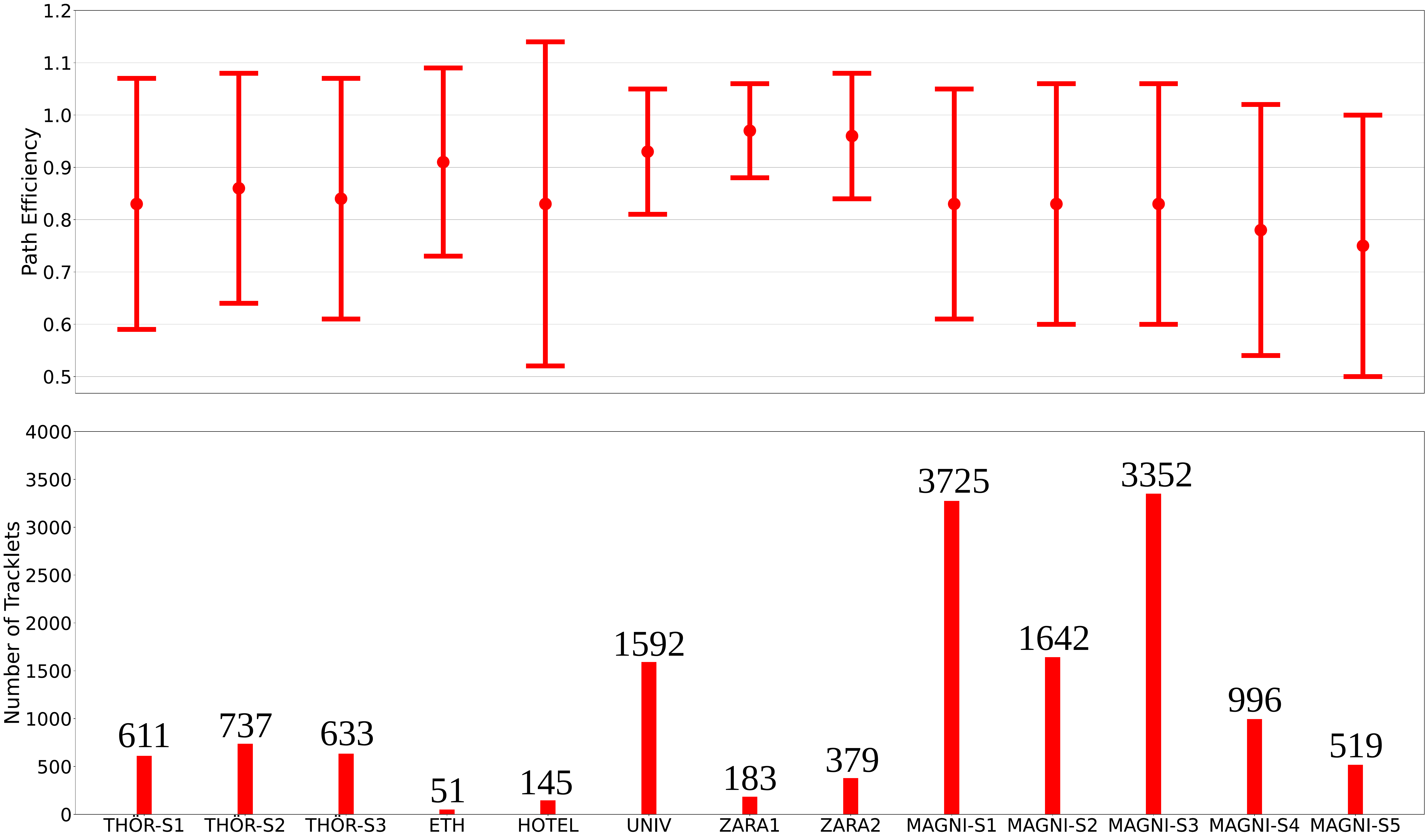}
\caption{\textbf{Top}: Path efficiency (mean ± one standard deviation) across datasets where lower results mean more linear trajectories. \textbf{Bottom}: Number of non-overlapping 8-second tracklets per dataset. TH\"{O}R-MAGNI provides the highest amount of non-linear trajectories.}
\label{fig:path_eff}
\end{figure}

Finally, we show the comparison of path efficiency and number of tracklets in Figure~\ref{fig:path_eff}. In terms of trajectory linearity, Scenarios~1--3 are aligned with the TH\"{O}R and HOTEL datasets, while the other datasets from the ETH/UCY benchmark contain more linear and less complex trajectories.  
It is also worth noting that TH\"{O}R-MAGNI Scenario~4 and 5 display the lowest average metrics 
($0.78$ and $0.75$, respectively). The presence of a moving robot might influence these scenarios, 
prompting human agents to navigate cautiously and align their motion with the robot's motion profile. Furthermore, TH\"{O}R-MAGNI presents a much higher number of non-overlapping tracklets than the other datasets.

In summary, these distinctive features make our dataset 
uniquely challenging, diverse, and valuable as a benchmark for evaluating human trajectory prediction 
methods. The heightened complexity and diverse range of trajectories in TH\"{O}R-MAGNI can provide a robust 
platform for evaluating the effectiveness of trajectory prediction methods, thereby increasing the breadth and depth of research in this area.

\section{Conclusions}\label{sec:conclusions}

In this paper, we present THÖR-MAGNI, a comprehensive dataset of human and robot navigation and interaction, extending THÖR~\citep{thor_20} with 3.5 times more motion data, novel interactive scenarios, and rich contextual annotations. Both datasets are accessible online at \url{http://thor.oru.se/}. To further support researchers, THÖR-MAGNI comes with a dedicated set of user-friendly tools --- a dashboard and a specialized Python package called \textit{thor-magni-tools} --- specifically designed to streamline the visualization, filtering, and preprocessing of raw data. These resources  aim to improve the accessibility and usability of the THÖR-MAGNI dataset. 

THÖR-MAGNI was created to fill a gap in datasets for human motion analysis, that was limiting HRI research: a lack of comprehensive inclusion of exogenous factors and essential target agent cues, which hinders holistic studies of human motion dynamics. Unlike existing datasets, THÖR-MAGNI includes a broader set of contextual features and offers multiple variations to facilitate factor isolation. Our dataset integrates different modalities, such as walking trajectories, eye tracking data, and environmental sensory inputs captured by a mobile robot. 

THÖR-MAGNI provides a comprehensive representation of diverse navigation styles of mobile robots and humans in shared environments, using multi-modal data. Our dataset contributes to the evolving landscape of human motion research through a comparative analysis with state-of-the-art datasets. Furthermore, we discuss the features of our dataset in the context of human motion and robot interaction, highlighting their importance in addressing gaps in the existing literature.
The THÖR-MAGNI dataset has already been used in research papers, demonstrating its usefulness for training role-conditioned motion prediction models~\citep{de_Almeida_2023_ICCV} and investigating  visual attention during human-robot interaction~\citep{schreiter2023advantages}.

In the future, we intend to propose a benchmark for multi-modal indoor trajectory prediction methods that builds on the rich contextual cues in TH\"{O}R-MAGNI. This effort is geared towards advancing the field by enabling the creation of more accurate models of human motion.

\begin{acks}
Authors would like to thank Johannes A. Stork for valuable
feedback and suggestions. The authors also thank all the AASS colleagues who helped to prepare and test the experimental infrastructure.
\end{acks}

\begin{funding}
This work was supported by the Wallenberg AI, Autonomous Systems and Software Program (WASP) funded by the Knut and Alice Wallenberg Foundation and by the European Union’s Horizon 2020 research and innovation program under grant agreement No. 101017274 (DARKO).
\end{funding}

\bibliographystyle{SageH.bst}
\bibliography{ref.bib}
\newpage
\appendix

\begin{table*}[!htbp]
\centering
\begin{tabular}{l|l|l}
\hline
\textbf{File ID} &
  \textbf{\begin{tabular}[c]{@{}l@{}}Visitors\\ Helmet ID 1-10\end{tabular}} &
  \textbf{\begin{tabular}[c]{@{}l@{}}Eyetrackers\\ Helmet ID 1-10\end{tabular}} \\ \hline
120522\_SC1A\_1 & 1,7,10 + 2,6 + 5 + 4 + 3 + 8 & 4 + 6 + 10 \\ \hline
120522\_SC1A\_2 & 8,4,3 + 10,5 + 1 + 2 + 6 + 7 & 4 + 6 + 10 \\ \hline
130522\_SC1A\_1 & 1,5,10 + 4,8 + 6 + 3 & 4 + 6 + 10 \\ \hline
130522\_SC1A\_2 & 5,6,8 + 3,10 + 1 + 4 & 4 + 6 + 10 \\ \hline
170522\_SC1A\_1 & 5,8 + 2,6 + 1 + 4 + 10 & 5 + 6 \\ \hline
170522\_SC1A\_2 & 1,4 + 5,10 + 2 + 6+ 8 & 5 + 6 \\ \hline
180522\_SC1A\_1 & 5,6 + 2,10 + 1 + 4+ 7 & 4 + 5+ 10 \\ \hline
180522\_SC1A\_2 & 1,4 + 6,10 + 2 + 5 + 7 & 4 + 5+ 10 \\ \hline
120522\_SC1B\_1 & 2,5,10 + 4,7 + 1 + 3 + 6 + 8 & 4 + 6 + 10 \\ \hline
120522\_SC1B\_2 & 3,6,7 + 1,8 + 2 +4 + 5 + 10 & 4 + 6 + 10 \\ \hline
130522\_SC1B\_1 & 1,5,10 + 4,8 + 3 + 6 & 4 + 6 + 10 \\ \hline
130522\_SC1B\_2 & 5,6,8 + 3,10 + 1 + 4 & 4 + 6 + 10 \\ \hline
170522\_SC1B\_1 & 1,6 + 2,5 + 4 + 8 + 10 & 5 + 6 \\ \hline
170522\_SC1B\_2 & 1,5 + 6,8 + 2 + 4 + 10 & 5 + 6 \\ \hline
180522\_SC1B\_1 & 2,6 + 7,10 + 1 + 4 +5 & 4 + 5+ 10 \\ \hline
180522\_SC1B\_2 & 1,6 + 4,5 + 2 + 7 + 10 & 4 + 5+ 10 \\ \hline
300922\_SC4A\_1 & 3 + 8 + 9 + 10   & 9 + 10 \\ \hline
300922\_SC4A\_2 & 3 + 8 + 9 + 10   & 9 + 10 \\ \hline
300922\_SC4A\_3  & 3,10 + 1 + 6 + 8 & 8 + 10 \\ \hline
300922\_SC4A\_4  & 3,10 + 1 + 6 + 8 & 8 + 10 \\ \hline
300922\_SC4B\_3 & 1,6 + 3 + 8 + 10 & 8 + 10 \\ \hline
300922\_SC4B\_4 & 1,6 + 3 + 8 + 10 & 8 + 10 \\ \hline
300922\_SC4B\_1  & 3 + 8 + 9 + 10   & 9 + 10 \\ \hline
300922\_SC4B\_2  & 3 + 8 + 9 + 10   & 9 + 10 \\ \hline
\end{tabular}
\caption{Assignment of \textbf{Visitors} and eyetrackers to helmet IDs in Scenarios~1 and 4. 
Each row relates the file ID to the assignment of \textbf{Visitors} roles (alone and groups), eyetrackers, 
and corresponding helmets IDs (1-10). The file ID contains ``date of the recording'' + "\_" + ``SC`` + 
``Scenario number`` + ``condition'' + "\_" + ``run number''.}
\label{tab:scenario1}
\end{table*}

\begin{table*}[!htbp]
\centering
\begin{tabular}{l|l|l|l|l}
\hline
\textbf{File ID} &
  \textbf{\begin{tabular}[c]{@{}l@{}}Visitors\\ ID 1-10\end{tabular}} &
  \textbf{\begin{tabular}[c]{@{}l@{}}Carrier\\ Box + Bucket\end{tabular}} &
  \textbf{\begin{tabular}[c]{@{}l@{}}Large Object\\ Leader + Follower\end{tabular}} &
  \textbf{\begin{tabular}[c]{@{}l@{}}Eyetracker\\ ID 1-10\end{tabular}} \\ \hline
120522\_SC2\_1 & 2,4,5+3+8   & 7+6  & 10+1 & 4+6+10 \\ \hline
120522\_SC2\_2&4,8+2+3+5&6+7&1+10&4+6+10\\ \hline
130522\_SC2\_1&1,6+3&8+10&5+4&4+6+10\\ \hline
130522\_SC2\_2&1+3+4&10+8&6+5&4+6+10\\ \hline
170522\_SC2\_1&2+5+8&4+10&6+1&4+5+6\\ \hline
170522\_SC2\_2&2,8+5&4+10&1+6&4+5+6\\ \hline
180522\_SC2\_1&1+5+6&10+2&7+4&4+5+10\\ \hline
180522\_SC2\_2&1,6+5&2+10&4+7&4+5+10\\ \hline
120522\_SC3A\_1&3,6,7+4+5&10+1&2+8&4+6+10\\ \hline
120522\_SC3A\_2&2,4,5+3+7&1+10&8+6&4+6+10\\ \hline
130522\_SC3A\_1&3,8+4&5+6&10+1&4+6+10\\ \hline
130522\_SC3A\_2&3,4+8&6+5&1+10&4+6+10\\ \hline
170522\_SC3A\_1&1,4+10&6+2&8+5&4+5+6\\ \hline
170522\_SC3A\_2&4,10+1&2+6&5+8&4+5+6\\ \hline
180522\_SC3A\_1&2+4+7&5+6&10+1&4+6+10\\ \hline
180522\_SC3A\_2&4,6+2&7+5&1+10&4+6+10\\ \hline
120522\_SC3B\_1&3,4,8+2+5&10+1&6+7&4+6+10\\ \hline
120522\_SC3B\_2&3,6,8+1+2&4+5&10+7&4+6+10\\ \hline
130522\_SC3B\_1&3,6+1&10+8&4+5&4+6+10\\ \hline
130522\_SC3B\_2&1,8+10&6+5&3+4&4+6+10\\ \hline
170522\_SC3B\_1&6,10+8&1+5&2+4&4+5+6\\ \hline
170522\_SC3B\_2&8,10+6&5+1&4+2&4+5+6\\ \hline
180522\_SC3B\_1&2,10+4&6+1&5+7&4+5+10\\ \hline
180522\_SC3B\_2&2,4,6&10+1&7+5&4+5+10\\ \hline
\end{tabular}
\caption{Assignment of \textbf{Visitors} and \textbf{Carriers} and eyetrackers to helmet ID in Scenarios~2 and 3. Each row relates the file ID to the assignment of the roles in Scenarios~2 and 3: {\em Visitors--Alone}, {\em Visitors--Groups 2}, {\em Visitors--Groups 3}, {\em Carrier--Box}, {\em Carrier--Bucket}, and {\em Carrier--Large Object}. The file ID contains ``date of the recording'' + "\_" + ``SC`` + ``Scenario number`` + ``condition'' + "\_" + ``run number''.}
\label{tab:scenario3}
\end{table*}

\begin{table}[!htbp]
  \centering
  \begin{tabular}{l|l|l|l}
  \hline
  \textbf{File ID} &
    \textbf{\begin{tabular}[c]{@{}l@{}}Visitors\\ ID 1-10\end{tabular}} &
    \textbf{\begin{tabular}[c]{@{}l@{}}Carrier\\ Storage Bin\end{tabular}} &
    \textbf{\begin{tabular}[c]{@{}l@{}}Eyetracker\\ ID 1-10\end{tabular}} \\ \hline
  300922\_SC5\_1   & 3 + 8 + 9 &  10        & 9 + 10 \\ \hline
  300922\_SC5\_2   & 3 + 8 + 10 &  9      & 9 + 10 \\ \hline
  300922\_SC5\_3   & 1,8 + 3 + 10 &  6     & 8 + 10 \\ \hline
  300922\_SC5\_4   & 1 + 3 + 6 + 8 & 10    & 8 + 10 \\ \hline

  \end{tabular}
  \caption{Assignment of human roles and eyetrackers to helmet IDs in Scenario~5. 
Each row relates the file ID to the assignment of \textbf{Visitors} roles (alone and groups of 2), {\em Carrier--Storage Bin HRI} role, eyetrackers, 
and corresponding helmets IDs (1-10). The file ID contains ``date of the recording'' + "\_" + ``SC`` + 
``Scenario number`` + ``condition'' + "\_" + ``run number''.}
  \label{tab:scenario5}
\end{table}

\begin{table*}[!htbp]
\centering
\begin{tabular}{l|l}
\hline
{ \textbf{Line of the header}}                                              & { \textbf{Description}}                                                    \\ \hline
FILE\textunderscore ID                                                         & Name of the file                               \\ \hline
N\textunderscore FRAMES\textunderscore QTM                                     & Amount of frames recorded at \SI{100}{Hz}                                         \\ \hline
N\textunderscore BODIES                                                        & Amount of rigid bodies present               \\ \hline
N\textunderscore MARKERS                                                       & Total amount of markers present                                      \\ \hline
CONTIGUOUS\textunderscore ROTATION\textunderscore MATRIX                       & Order of a 3x3 rotation matrix         \\ \hline
\begin{tabular}[c]{@{}l@{}}MODALITIES\textunderscore\\ WITH\textunderscore UNITS\end{tabular}                            & List of measured variables               \\ \hline
\begin{tabular}[c]{@{}l@{}}MODALITIES\textunderscore\\ UNITS\textunderscore SPECIFIED\end{tabular}                        & Measurement units specified                                                      \\ \hline
\begin{tabular}[c]{@{}l@{}}EYETRACKING\textunderscore\\ DEVICES\end{tabular}  &
  \begin{tabular}[c]{@{}l@{}}List of eye trackers in this recording \\ (Not all devices are used in all files)\end{tabular} \\ \hline
\begin{tabular}[c]{@{}l@{}}EYETRACKING\textunderscore\\ FREQUENCY\textunderscore IR\end{tabular}                           & Frequency of eyetrackers infra red sensor         \\ \hline
\begin{tabular}[c]{@{}l@{}}EYETRACKING\textunderscore\\ FREQUECNY\textunderscore SCENE\textunderscore CAMER\end{tabular} & Frequency of scene cameras video                                \\ \hline
\begin{tabular}[c]{@{}l@{}}EYETRACKING\textunderscore\\ DATA\textunderscore INCLUDED\end{tabular}                         & Data included from the eye tracker                    \\ \hline
EYETRACKING\textunderscore DATA\textunderscore N\textunderscore FRAMES &
  Total amount of frames with this data \\ \hline
BODY\textunderscore NAMES                                                      & Name of each rigid body\\ \hline
BODY\textunderscore ROLES                                                      & Role label of each rigid body                           \\ \hline
BODY\textunderscore NR\textunderscore MARKERS                                  & Amount of markers for each rigid body                                        \\ \hline
MARKER\textunderscore NAMES &
  \begin{tabular}[c]{@{}l@{}}The names of all markers used in this file\\ (Unaligned with the previous rows)\end{tabular} \\ \hline
\end{tabular}
\caption{Overview of the metadata in the headers of the CSV files. It includes recording information such as the file ID, date, scenario, condition, and run. It also provides details on data quantities, including the number of frames recorded, rigid bodies present, and markers used. Moreover, it includes information about the order of rotation matrices, measured variables with units, specified measurement units, and a list of eye trackers utilized in each recording. Furthermore, it outlines the frequencies of eyetracker sensors and scene cameras, as well as the presence of eyetracking data. Finally, it covers details about rigid bodies including their names, roles, and the number of markers associated with them.}
\label{tab:CSVHead}
\end{table*}

\end{document}